\documentclass[accepted]{melba}

\usepackage{mwe} 

\usepackage{dblfloatfix}
\usepackage{placeins}
\usepackage{booktabs}
\usepackage{enumitem}
\newcommand{\Diff}[0]{\mathrm{Diff}}    

\usepackage{xcolor}

\usepackage{amsmath}  
\usepackage{amssymb}  
\usepackage{bm}       
\usepackage{float}
\usepackage{makecell} 
\usepackage{multirow}
\usepackage{relsize}
\usepackage{algorithm} 
\usepackage{algorithm}
\usepackage{algpseudocode}
\algrenewcommand\algorithmicrequire{\textbf{Input:}}
\algrenewcommand\algorithmicensure{\textbf{Output:}}
\usepackage{natbib}
\usepackage{float}
\usepackage{caption}
\usepackage{graphicx}
\usepackage{wrapfig}    
\usepackage{amsmath,amsfonts}

\melbaid{2026:021}  
\doi{10.59275/j.melba.2026-eec1}
\melbaauthors{Nian Wu, Nivetha Jayakumar, Jiarui Xing, and Miaomiao Zhang}  
\email{bsw3ac@virginia.edu}
\volume{2026}
\firstpageno{440}  
\melbayear{2026}  
\datesubmitted{2025-11}  
\datepublished{2026-08}  

\melbaspecialissue{Medical Imaging with Deep Learning (MIDL) 2020}
\melbaspecialissueeditors{Marleen de Bruijne, Tal Arbel, Ismail Ben Ayed, Hervé Lombaert}

\ShortHeadings{Geodesic-informed Generative Diffusion Model}{Wu, Jayakumar, Xing, and Zhang}

\title{Geodesic-informed Generative Diffusion Model For Topology-preserved Image Video Generation}

\author{
	\firstname Nian \surname Wu\aff{1}\orcid{https://orcid.org/0000-0002-9168-3518},
    \firstname Nivetha \surname Jayakumar\aff{1}\orcid{https://orcid.org/orcid.org/0009-0009-7515-8447},
    \firstname Jiarui \surname Xing\aff{3}\orcid{https://orcid.org/0009-0005-4584-8536},
	\name Miaomiao \surname Zhang\aff{1,2}\orcid{https://orcid.org/0000-0003-0457-3335}
}

\affiliations{
	\num 1 \addr University of Virginia, Department of Electrical \& Computer Engineering, Charlottesville, VA, USA \\
	\num 2 \addr University of Virginia, Department of Computer Science, Charlottesville, VA, USA \\
    \num 3 \addr Yale School of Medicine, New Haven, CT, USA \\
}

\abstract{
	Generative diffusion models have emerged as a class of powerful techniques for various imaging  applications, including but not limited to synthesis, reconstruction, and segmentation. Despite their success, current generative models pose two key limitations. First, they primarily rely on image intensity and texture information, with limited attention to underlying object geometry. As a result, they do not guarantee geometric or topological consistency during the generation process, which is a crucial requirement for high-stakes domains such as computational anatomy, biology, and robotics, where preserving object structure is critical. Second, existing models fail to explicitly learn or represent shape changes in the generative process. Such deformation dynamics remain occluded within network parameters; hence leaving the transformation process uninterpretable and physically uninformed. To address these challenges, we introduce IGG ({\bf I}mage {\bf G}eneration informed by {\bf G}eodesic dynamics), a novel framework that integrates topology-preserving geodesic principles into the diffusion-based generative process. In contrast to conventional methods that operate in image intensity space, IGG learns and synthesizes diverse samples within geodesic deformation spaces, where geometric object changes are learned as smooth and invertible smooth mappings from a given template/source image. Specifically, IGG employs a two-stage architecture: (i) a geodesic-informed image registration (GIR) module that directly encodes geodesic paths of image deformations into a compact latent space, and (ii) a latent geodesic diffusion (LGD) model that captures the distribution of these deformation representations, conditioned on a template image and/or text prompts. We validate IGG on the datasets of plant growth and brain MRI scans. Experimental results demonstrate that IGG outperforms state-of-the-art image generation and editing models, producing realistic, high-quality images with preserved topology and fewer artifacts. Furthermore, when used for data augmentation in downstream segmentation tasks, IGG substantially improves segmentation accuracy, particularly in low-data regimes. Our code is publicly available at ~\url{https://github.com/nellie689/IGG}.
    }


\begin{document}

\twocolumn[\maketitle]

\section{Introduction}
\label{sec:intro}
Recent advances in deep generative modeling have significantly impacted various image analysis tasks, including but not limited to image synthesis~\citep{xing2024lamod,xie2025meddiff,liu2021generative,reinhold2021structural}, reconstruction~\citep{jayakumar2023sadir,tezcan2018mr}, translation~\citep{deb2025unsupervised,graf2023denoising,ozbey2023unsupervised}, and more~\citep{yang2019unsupervised,hossain2025mgaug}. Early research works, such as the variational autoencoder (VAE), assume a Gaussian prior over the latent feature space and learn its approximate posterior through variational inference~\citep{kingma2013auto}. Although effective, these models can produce overly smooth or blurry outputs due to their simplified assumptions~\citep{hossain2025mgaug}. Another category of generative modeling is the generative adversarial networks (GANs), which learn to produce realistic data samples by training a generator to compete against a discriminator that distinguishes generated samples from real data~\citep{goodfellow2020generative}. However, such models may suffer from mode collapses, where the generator produces limited diversity of outputs by ignoring other modes of the true data distribution~\citep{zhang2018convergence,bang2021mggan,li2021tackling}. In contrast, diffusion models have recently emerged as a powerful class of deep generative frameworks that achieve state-of-the-art performance in modeling complex, multimodal data distributions and generating photorealistic images across diverse applications~\citep{mukhopadhyay2023diffusion,ho2020denoising,deb2025unsupervised}. These models employ an implicit generative scheme by learning to reverse a gradual noising process and iteratively transforming random noise into fine-grained and high-fidelity image samples~\citep{ho2020denoising}. Due to these capabilities, diffusion models have demonstrated remarkable success in synthesizing highly realistic medical images to address critical challenges in biomedical research, such as data/label scarcity~\cite{sharma2023medic,jayakumar2024tpie}, missing or incomplete modalities~\cite{jayakumar2023sadir,deb2025unsupervised}, and the need for predictive imaging to support disease progression modeling and clinical decision-making~\cite{yoon2023sadm,dao2024conditional,zhang2024development}.

Despite the success of the aforementioned generative diffusion models, current approaches face several limitations that can constrain their practical use in high-stakes domains requiring high accuracy and fidelity, such as medical imaging, robotics, and autonomous systems. In particular, most existing diffusion models focus on image intensity and texture, with little explicit consideration of object geometry during learning and generation~\citep{nguyen2024visual,li2024blip,hertz2022prompttoprompt,kawar2023imagic}. As a result, they may produce unrealistic or biologically implausible samples, which poses significant risks for downstream applications where accurate geometric preservation is critical~\citep{azizi2023synthetic,jayakumar2023sadir,starck2025diff}. Recent advances have introduced simple geometric constraints such as shape localization~\citep{patashnik2023localizing,gupta2024topodiffusionnet}, boundary conditions~\citep{maze2023diffusion}, and 3D shape priors~\citep{yu2025surf,hu2024topology} in the modeling process. However, these approaches lack fine-grained structural preservation that is highly desirable in sensitive domains, including computational anatomy and medical imaging. Another important yet under-investigated limitation of current diffusion models is their inability to provide interpretable metrics or quantitative measures of topological integrity for image objects. Commonly used evaluation metrics include Fréchet Inception Distance (FID) for assessing distribution similarity between real and generated data~\citep{heusel2017gans}, Inception Score (IS) for measuring diversity and quality~\citep{salimans2016improved}, and Structural Similarity Index (SSIM) for evaluating pixel-level fidelity~\citep{wang2004image}. While these metrics effectively assess visual quality, they do not account for the preservation of object geometry and topology, raising questions about their validity in structure-sensitive applications.

In addition, existing diffusion models do not explicitly capture or represent object shape changes during the generative process~\citep{ho2020denoising,dao2024conditional,jayakumar2024tpie,deb2025unsupervised,yoon2023sadm}. Recent studies have demonstrated that incorporating a geodesic-constrained deformation process to model object shape changes within the network can promote physically interpretable deformation trajectories, which enables the registration framework to capture the underlying dynamics of shape evolution rather than performing purely data-driven alignment~\citep{wu2023neurepdiff,wu2024learning}. Moreover, this geodesic guidance provides an additional source of regularization in the deformation learning process; thereby improving the biological plausibility of the registration results by suppressing artifacts such as folding, tearing, and crossing in the predicted transformations.

This paper presents an extended work of {\em Image Generation informed by Geodesic dynamics (IGG)}~\citep{wu2025igg}, that for the first time generates images by deforming a given template along random geodesics in deformation spaces guided by text instructions. In contrast to current approaches~\citep{brooks2023instructpix2pix,ho2022video,kim2022diffusemorph}, the IGG framework ensures topological consistency by treating each generated image as a deformed variation of a template or reference image through learned diffeomorphic transformations (also known as a one-to-one, smooth, and invertible mapping). A dynamic geodesic, producing natural transformations and providing interpretable metrics to quantify topological changes, are explicitly learned during the diffusion process. More specifically, the IGG model comprises two key components: a latent representation learning framework of diffeomorphic transformations informed by geodesic dynamics, and a latent geometric diffusion model conditioned on user-defined text instructions. Building on top of IGG~\citep{wu2025igg}, our work further 
\begin{enumerate}[label=(\roman*)]
\item Expand the core method section of IGG by introducing details of the geodesic-informed image registration module, which provides a more principled representation of topology-preserving transformations.
\item Thoroughly investigate the quantitative metrics to assess both the fidelity and topological consistency of the generated samples.
\item Design downstream experiments in the context of image segmentation to demonstrate an additional clinical utility of the generated samples in real-world applications.  
\end{enumerate}

We demonstrate the effectiveness of IGG on a diverse set of real-world datasets, including Komatsuna plant growth data~\citep{uchiyama2017easy} and brain MRIs~\citep{lamontagne2019oasis}. Experimental results show that IGG significantly outperforms state-of-the-art text-guided generative models~\citep{ho2022video,yang2024cogvideox,xing2025dynamicrafter} in producing image samples with well-preserved topological structures. We further demonstrate the potential of IGG to enhance deep learning-based downstream tasks by training brain segmentation networks with synthesized image-label pairs. It is worth noting that, once trained, IGG generalizes well across image domains (e.g., variations in image contrast and scanner types) due to its flexibility in generating geometric deformations conditioned on any given template.

\section{Background: Geodesics In Deformation Spaces}
\label{sec:backgroundlddmm}
In this section, we briefly review the basic concept of geodesics in deformation spaces, which provides a principled way to model smooth and natural deformations between images while preserving structural integrity~\citep{miller2002metrics,beg2005computing}. With the underlying assumption that objects of a generic class (e.g., human brains, hearts, or lungs) are described as deformed variants of a given template, descriptors of that class arise naturally by deforming the template to other images along a {\bf geodesic} - a shortest path with locally minimized energy of transformations~\citep{avants2008symmetric,joshi2004unbiased}. 
In theory, every topological property of the deformed template can be preserved by enforcing the resulting transformations to be diffeomorphisms, i.e., differentiable, bijective mappings with differentiable inverses~\citep{beg2005computing,arnold1966,miller2002metrics}. Violations of such constraints introduce image artifacts, such as tearing, crossing, or passing through itself. 

\begin{figure}[t]
    \centering
    \includegraphics[width=\linewidth]{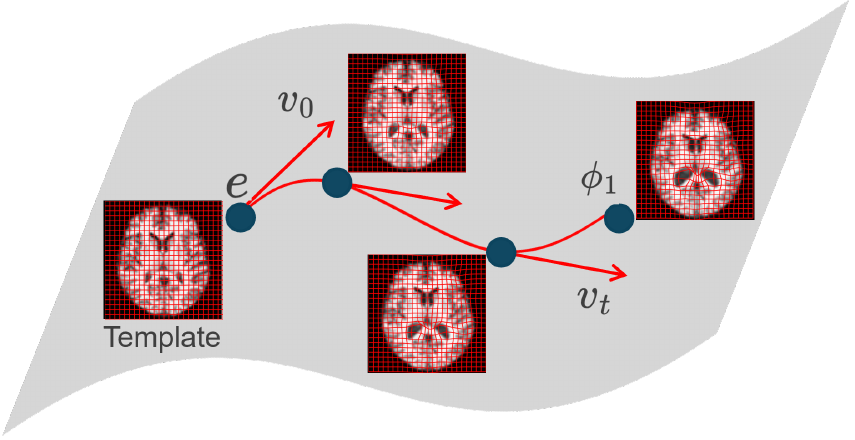}
    \caption{An illustrated example of a geodesic path in the space of brain deformations.}
    \label{fig:intro}
\end{figure}

Let $\Diff^\infty(\Omega)$ denote the space of smooth
diffeomorphisms on a $d$-dimensional torus domain $\Omega = \mathbb{R}^d / \mathbb{Z}^d$. The tangent space of diffeomorphisms is the
space $V = \mathfrak{X}^\infty(T\Omega)$ of smooth vector fields on $\Omega$. Consider a time-varying velocity field, $\{v_t\} : [0, 1] \rightarrow V$, we can generate diffeomorphisms $\{\phi_{t}\}$ between pairwise images by solving
\begin{equation}
\label{eq:phi_v}
\frac{d \phi_t}{dt} = v_t(\phi_t), \, t \in [0, 1], 
\end{equation}
where $\phi_0$ is the identity map, $e$, and $\phi_1$ is the target transformation. 

The geodesic minimizes the functional $\int_0^1 (\mathcal{L} v_t, v_t) \, dt$~\citep{beg2005computing}, where $\mathcal{L}: V\rightarrow V^{*}$ is a symmetric, positive-definite differential operator that maps a tangent vector $ v(t)\in V$ into its dual space as a momentum vector $m(t) \in V^*$. We typically write $m(t) = \mathcal{L} v(t)$, or $v(t) = \mathcal{K} m(t)$, with $\mathcal{K}$ being an inverse operator of $\mathcal{L}$. In this paper, we adopt a commonly used Laplacian operator $\mathcal{L}=(- \alpha \Delta + e)^3$, where $\alpha$ is a weighting parameter that controls the smoothness of transformation fields. The $(\cdot, \cdot)$ is a dual pairing, which is similar to an inner product between vectors. An example of geodesic path in the space of deformations is illustrated in Fig.~\ref{fig:intro}.

According to a well-known geodesic shooting algorithm~\citep{vialard2012}, the minimum of the functional mentioned above is uniquely determined by solving a Euler-Poincar\'{e} differential (EPDiff) equation~\citep{vialard2012,younes2009evolutions} with a given initial condition. That is, for $\forall v_0 \in V$, a geodesic path $t \mapsto \phi_t$ in the space of diffeomorphisms can be computed by forward shooting the EPDiff equation
\begin{equation}
\label{eq:epdiff}
    \frac{\partial v_t}{\partial t} =-K\left[(Dv_t)^Tm_t + Dm_t\, v_t + m_t \operatorname{div} v_t\right],
\end{equation}
where $D$ denotes the Jacobian matrix and $\operatorname{div}$ is a divergence operator.

\paragraph*{Derive geodesics from images.} Consider a source image $S$ and a target image $T$ defined in the domain $\Omega$ ($S(x), T(x):\Omega \rightarrow \mathbb{R}$). The optimization of geodesic transformations can be formulated as minimizing an energy function over the initial velocity, subject to the EPDiff equations, i.e., 
\begin{equation*}
\label{eq:lddmm}
 E(v_0) = (\mathcal{L} v_0, v_0)  + \lambda  \text{Dist}(S(\phi_1), T)  \, \, \text{s.t. Eq.}~\eqref{eq:phi_v}  \, \& \, ~\eqref{eq:epdiff}.
\end{equation*}
Here, Dist(·,·) is a distance function that measures the dissimilarity between images, and $\lambda$ is a positive weighting parameter. In this paper, we will use the commonly used sum-of-squared intensity differences~\citep{beg2005computing,wu2024tlrn,zhang2017frequency}. 

\section{Method: IGG}
\begin{figure*}[!t]
\centering
\includegraphics[width=\textwidth]{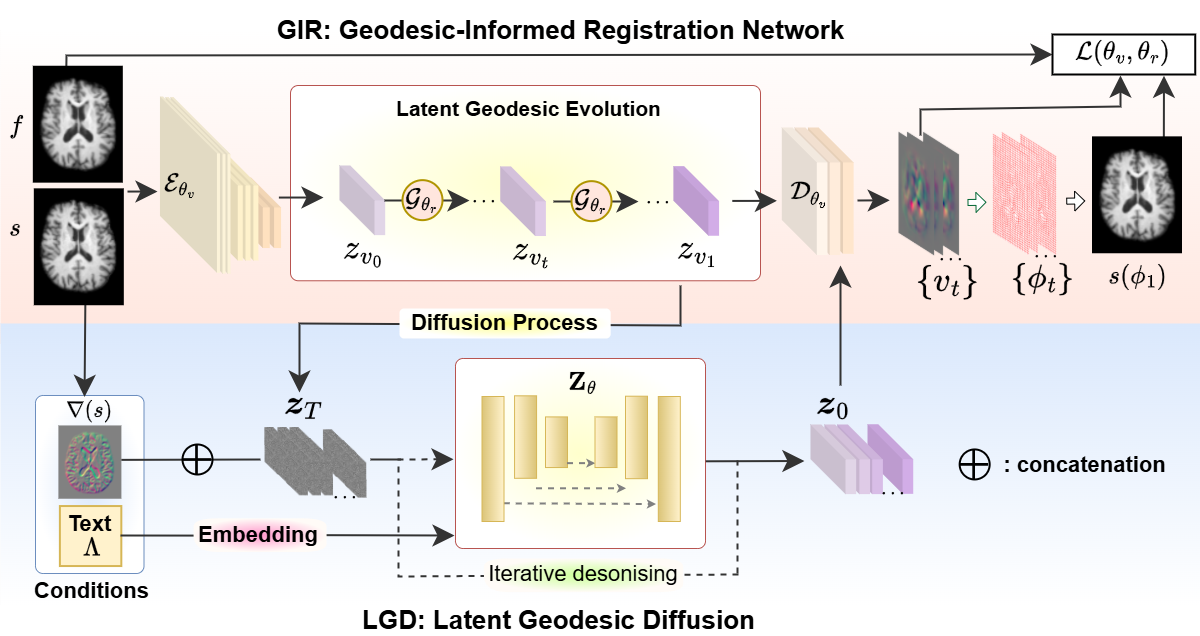}
\caption{An overview of our IGG framework with two key components: (i) GIR to learn the latent representations of geodesic transformations from paired template and target images; and (ii) LGD to generate a time-sequence of image transformations over time, conditioned on text instructions.}
\label{fig:Arc}
\end{figure*}

This section introduces an extended work of, IGG, that for the first time generates images in deformation spaces guided by text instructions~\citep{wu2025igg}. A dynamic geodesic, producing natural transformations and providing interpretable metrics to quantify topological changes, will be explicitly learned during the generative diffusion process. Our IGG model consists of two main components: (i) a geodesic-informed image registration (GIR) network that directly encodes geodesic paths of image deformations into a compact latent space; and (ii) a latent geodesic diffusion (LGD) model that captures the latent distribution of
 these deformation representations, conditioned on a template image and/or user-defined text prompts. An overview of our model is illustrated in Fig.~\ref{fig:Arc}. 

\subsection{Representation Learning of Geodesic Deformations via GIR}
Consider a given set of $N$ template/source and target image pairs associated with text instructions $\{s^n, f^n, \Lambda^n\}_{n=1}^N$. We introduce a GIR network, based on an autoencoder architecture, to learn the latent representation of geodesic transformations, parameterized by velocity fields $\{v_t\}$, between a template $s^n$ and a target $f^n$. 
The encoder of GIR, $\mathcal{E}_{\theta_v}$, maps input data to the latent representation of a sequence of velocity fields, $\{z_{v_t}\}_{t=0}^{1}$. Inspired by the recent work NeurEPDiff~\citep{wu2023neurepdiff}, which develops a neural operator to learn mappings governed by the EPDiff equation, we incorporate the NeurEPDiff module, $\mathcal{G}_{\theta_r}$, into our IGG model to learn geodesics in latent deformation spaces. More specifically, starting with a latent initial velocity $z_{v_0}$, the operator $\mathcal{G}_{\theta_r}$ will iteratively propagate it along the geodesic path, $z_{v_0} \longmapsto \cdots z_{v_t} \longmapsto \cdots z_{v_1}$, through a designed latent geodesic evolution module. 

\paragraph*{Latent Geodesic Evolution.} Similar to~\citep{wu2023neurepdiff,wu2024learning}, we leverage neural operators as surrogate models to approximate the mapping function of the EPDiff equation during network training. In particular, we employ a multilayer neural network with $J$ hidden layers to simulate the geodesic mapping function (EPDiff) from $z_{v_{t}}$ to $z_{v_{t+1}}$. For notational simplicity, we let $u^j$ denote the input of the $j$-th hidden layer and $u^{j+1}$ its output, where $j \in \{1, \cdots, J\}$. At each hidden layer, we combine a \textit{local linear transformation} $W^j$ with a \textit{global convolutional kernel} $\mathcal{H}^j$, enabling the extraction of both \textit{local} and \textit{global representations}. 
A nonlinear activation function, $\sigma(\cdot)$, is defined as a Gaussian Error Linear Unit (GeLU)~\citep{hendrycks2016gaussian} with a smoothing operator $K$ to ensure continuity and smoothness in the output signal, i.e., $\sigma(\cdot) := K (\text{GeLU} (\cdot))$. Formally, we update $u^{j+1} \leftarrow u^{j}$ by
\begin{align}
   u^{j+1} := \sigma(W^j u^{j} + \mathcal{H}^j*u^{j}),
   \nonumber
\end{align}
where $*$ represents a convolution operator. Note that, for computational efficiency, the global convolution is implemented as an element-wise multiplication in the Fourier domain, followed by an inverse transform back to the spatial domain.  

Finally, the geodesic evolution of the latent velocity from $z_{v_{t}}$ to $z_{v_{t+1}}$ at each time point $t$ can be defined as
\begin{equation*}
z_{v_{t+1}}:= \sigma(W^J, \mathcal{H}^J) \circ \cdots \circ \sigma(W^2, \mathcal{H}^2) \circ \sigma(W^1, \mathcal{H}^1, z_{v_t}),
\label{eq:FneurEPDiff}
\end{equation*}
where $\circ$ denotes the composition of network operations. The decoder, $\mathcal{D}_{\theta_v}$, then projects these latent representations back to the full-dimensional image space to produce a geodesic flow of velocity fields, $v_0 \longmapsto \cdots v_t \longmapsto \cdots v_1$. The corresponding deformations, $\phi_0 \longmapsto \cdots \phi_t \longmapsto \cdots \phi_1$ are then generated according to Eq.~\ref{eq:phi_v}. For simplicity of notation, we define $\boldsymbol{\Phi} \overset{\Delta}{=} \{\phi_t\}_{t=0}^1$ in the following sections. 

The network loss combines contributions from an unsupervised registration loss, and a geodesic loss guided by numerical solutions of the EPDiff equation obtained via Euler integration. These solutions are represented as $\{\hat{v}_t^n \}, t \in (0, 1]$ with a simultaneously learned initial velocity $\hat{v}_{0}^n$. To simplify the notation, we define $\bm{\hat{v}}^n \triangleq \{ \hat{v}_t^n \}$ before formulating the loss function as
\begin{align}
\label{eq:JointLossFunGeodesic}
\mathcal{L}(\theta_v, \theta_r) &=  \sum_{n=1}^{N} \lambda \, \|(s^n(\phi^{n}_1 (\theta_v)) - f^n \|^2_2  \nonumber \\
&+ \frac{1}{2} (\mathcal{L} v^n_0 (\theta_v), v^n_0 (\theta_v))  \nonumber \\
&+ \eta \, \| \mathcal{D}_{\theta_v} (\mathcal{G}_{\theta_r}(z_{v_0}))- \bm{\hat{v}}^n \|_2^2  \nonumber \\
&+ \text{Reg}(\theta_r, \theta_v),
\,\,\,\, s.t. \,\, \text{Eq.} ~\eqref{eq:phi_v}, 
\end{align}
where $\lambda$ and $\eta$ are positive weighting parameters to balance the image matching term and the geodesic loss, and $\text{Reg}(\cdot)$ is a network regularization.

\subsection{Latent Geodesic Diffusion Model}
With a pre-trained GIR network, we are now ready to introduce a latent geodesic diffusion module to simulate the distribution of latent geodesic flows for image transformations, \(\{ z^n_{v_t}\}_{t=0}^{1} \). To simplify the notation, we define $\boldsymbol{z}^{n} := \{ z^n_{v_t}\}_{t=0}^{1}$ for the $n$-th subject, where all time-dependent components $\{z^n_{v_t}\}$ are concatenated to form a unified representation. In contrast to previous approaches that perform the diffusion process in the image space~\citep{brooks2023instructpix2pix,ho2022video,kim2022diffusemorph}, our latent geometric diffusion operates in the deformation space. Specifically, our diffusion process directly samples geodesics of topology-preserving diffeomorphic transformations, which are then applied to deform the input template to generate the final output.

\subsubsection{Forward diffusion in geodesic deformation spaces.}
Following the principles of denoising diffusion probabilistic models (DDPM)~\citep{ho2020denoising} and video diffusion models~\citep{ho2022video}, our forward diffusion is fixed to a Markov chain that progressively adds noise, $\epsilon \sim \mathcal{N}(0, \mathbf{I})$, to the initial latent geodesic representations $\boldsymbol{z}_0^n \sim p(\boldsymbol{z}_0)$, with a number of $T$ steps. Mathematically, the forward diffusion process can be formulated as
\begin{equation}
q(\boldsymbol{z}_{\tau}^n \mid \boldsymbol{z}_{\tau-1}^n) := \mathcal{N}\left(\boldsymbol{z}_{\tau}^n; \sqrt{1-\beta_{\tau}}\,\boldsymbol{z}_{\tau-1}^n, \beta_{\tau} \mathbf{I}\right),
\nonumber
\end{equation}
where $\tau \in [1, \cdots, T]$ and $\beta_\tau$ is a time-dependent variance schedule of scalar values $\in (0, 0.999]$ used to parameterize the probabilistic transitions. Alternatively, the forward diffusion process of sampling $\boldsymbol{z_{\tau} }$ at an arbitrary timestep $\tau$ can be formulated in closed form using
\begin{align}
    q(\boldsymbol{z}_{\tau}^n | \boldsymbol{z}_{0}^n) &= \mathcal{N}(\boldsymbol{z}_{\tau}^n; \sqrt{\bar{\alpha}_\tau}\boldsymbol{z}_{0}^n, (1-\bar{\alpha}_\tau)\mathbf{I}), \nonumber \\
    \boldsymbol{z}_{\tau}^n &= \sqrt{\bar{\alpha}_\tau}\boldsymbol{z}_{0}^n + \sqrt{1-\bar{\alpha}_\tau} \epsilon, \label{eq:dif-forward-onestep-det}
\end{align}
where $\alpha_\tau = 1- \beta_\tau$ and $\bar{\alpha}_\tau = \prod_{s=1}^\tau \alpha_s$, where $s \in [1, \cdots, \tau]$.

\subsubsection{Reverse diffusion with geometric conditioning.}
Similar to \cite{ho2020denoising}, the reverse diffusion process iteratively denoises the latent variable of the diffusion model $\boldsymbol{z}_{\tau}^n$ over $\tau$ steps, given template-specific conditions. We introduce a novel geometric conditioning in the latent geodesic space, which enables the model to adapt more effectively to specific contexts; hence improving its generative performance based on observed template images and provided text instructions. More specifically, our geometric conditioning is achieved by concatenating two components: image embedding in the form of downsampled template image gradient, $\{\nabla s^n\}$ and the geodesic deformation embedding, represented by the latent velocity fields, $\{\boldsymbol{z}_{\tau}^n\}$. Following a similar principle as seen in~\citep{ho2020denoising}, we sample the denoised latent geodesic flow of velocity fields $\boldsymbol{z}^n_0$, starting from the noisy state $\boldsymbol{z}^n_{\tau}$ at the $\tau$-th time-step and progressively remove noise to reconstruct the original latent representation. This process is mathematically expressed as
\begin{equation}
\begin{gathered}
    p_\theta(\boldsymbol{z}^{n}_{\tau-1} | \boldsymbol{z}^{n}_{\tau}, \nabla s^{n} , \Lambda^{n}, \tau) = \nonumber \\
    \mathcal{N}(\boldsymbol{z}^{n}_{\tau-1}; \mu_\theta(\boldsymbol{z}^{n}_{\tau}, \nabla s^{n} , \Lambda^{n}, \tau),\Sigma_\theta(\boldsymbol{z}^{n}_{\tau}, \nabla s^n , \Lambda^n, \tau)),
    \nonumber
\end{gathered}
\end{equation}

where $\mu_\theta$ and $\Sigma_\theta$ represent the parameterized mean and variance of the approximated Gaussian distribution at each time step of the reverse diffusion process~\citep{ho2022video} . 
In our experiments, we employ a 3D UNet architecture~\citep{ronneberger2015u} ${\mathbf{Z}_\theta}$ as the deep predictive model, which predicts the noise to be removed from $\boldsymbol{z}^{n}_{\tau}$ to reconstruct $\boldsymbol{z}^{n}_{\tau-1}$ using the equation:
\begin{gather}
    \boldsymbol{z}_{\tau-1}^n = \frac{1}{\sqrt{\alpha_\tau}} (\boldsymbol{z}_{\tau}^n - \frac{1-\alpha_\tau}{\sqrt{1-\bar{\alpha}_\tau}}{\mathbf{Z}}_{\theta}(\boldsymbol{z}^{n}_{\tau}, \nabla s^n , \Lambda^n, \tau))+ \sigma_t I,
    \label{eq:dif-back} 
\end{gather}
where $I \sim \mathcal{N}(0, \mathbf{I})$ and $\sigma_t = \tilde{\beta}_t = \frac{1 - \bar{\alpha}_{t-1}}{1 - \bar{\alpha}_t} \cdot \beta_t$.

The text embedding, encoded using a pre-trained CLIP text encoder~\citep{radford2021learning}, $\{\Lambda^n\}$ is integrated into this network via cross-attention. With an optimized denoising network ${\mathbf{Z}_\theta}$, the iterative sampling using Eq~\ref{eq:dif-back} yields a clean geodesic deformation embedding $\boldsymbol{z}_0^n$. 

The network loss of our proposed latent geodesic diffusion module is formulated as
\begin{equation}
\label{eq:IGG}
\mathcal{L}_{\theta} = \frac{1}{N}\mathlarger{\sum}_{n=1}^N \vert\vert \epsilon_{\tau}^n - {\mathbf{Z}}_{\theta}(\boldsymbol{z}^{n}_{\tau}, \nabla s^n , \Lambda^n, \tau) \vert\vert^2_2  + \text{reg}(\theta), \tau \in \text{U}[1,T],
\end{equation}
where $\text{reg}(\cdot)$ is a regularization of the network parameter $\theta$, U denotes a uniform  distribution and $T$ is the total number of diffusion timesteps.

{\bf Classifier-free guidance.} To balance the trade-off between sample quality and diversity of velocity fields in the latent space after training, we optimize the combination of conditional and unconditional diffusion models. This is achieved by leveraging image-conditioned guidance with scale $\delta_\mathrm{I}$, text-conditioned guidance with scale $\delta_\mathrm{T}$, and a null-condition $\varnothing$. Following similar principles from the classifier-free guidance approach~\citep{ho2022classifier}, we first jointly train a conditional and an unconditional IGG model. The resulting score estimates from these models are then combined to predict the noise $\hat{\mathbf{Z}}$ as
\begin{equation}
\begin{gathered}
\hat{{\mathbf{Z}}}_{\theta}(\boldsymbol{z}^{n}_{\tau}, \nabla s^n , \Lambda^n, \tau) \nonumber \\
= {\mathbf{Z}}_{\theta}(\boldsymbol{z}^{n}_{\tau}, \varnothing, \varnothing, \tau) \nonumber \\
+ 
 \delta_\mathrm{I} \cdot \left[{\mathbf{Z}}_{\theta}(\boldsymbol{z}^{n}_{\tau}, \nabla s^n , \varnothing, \tau) -  {\mathbf{Z}}_{\theta}(\boldsymbol{z}^{n}_{\tau}, \varnothing, \varnothing, \tau)\right] \nonumber \\ 
 \quad + \delta_\mathrm{T} \cdot \left[{\mathbf{Z}}_{\theta}(\boldsymbol{z}^{n}_{\tau}, \nabla s^n , \Lambda^n, \tau) - {\mathbf{Z}}_{\theta}(\boldsymbol{z}^{n}_{\tau}, \nabla s^n, \varnothing, \tau)\right] \nonumber \\ \nonumber \\ 
 = (1-\delta_\mathrm{I}) \cdot {\mathbf{Z}}_{\theta}(\boldsymbol{z}^{n}_{\tau}, \varnothing, \varnothing, \tau) \nonumber \\ + (\delta_\mathrm{I} - \delta_\mathrm{T}) \cdot {\mathbf{Z}}_{\theta}(\boldsymbol{z}^{n}_{\tau}, \nabla s^n , \varnothing, \tau) \nonumber \\ 
 \quad + \delta_\mathrm{T} \cdot {\mathbf{Z}}_{\theta}(\boldsymbol{z}^{n}_{\tau}, \nabla s^n , \Lambda^n, \tau).
 \nonumber
\end{gathered}
\end{equation}

\subsection{Network Optimization}  
We first optimize the loss of representation learning of geodesic transformations (Eq.~\eqref{eq:JointLossFunGeodesic}), followed by the latent geodesic diffusion model (Eq.~\eqref{eq:IGG}). The training process of IGG is summarized in Alg.~\ref{alg:training}. During testing, the procedure for sampling the geodesic of latent velocity fields, conditioned on a given template image and text instructions, is outlined in Alg.~\ref{alg:sampling}.

\begin{algorithm}[!h]
\caption{IGG Training}
\label{alg:training}
\begin{algorithmic}[1]
\Require Data $\{ s^n, f^n, \Lambda^n \}_{n=1}^{N_{\text{train}}}$

\State Pretrain $\mathcal{E}_{\theta_v}, \mathcal{E}_{\theta_r}, \mathcal{D}_{\theta_v}$ via Eq.~\eqref{eq:JointLossFunGeodesic}
\Repeat
    \State Initial latent geodesic representations, $\boldsymbol{z}^{n}_0 \gets \mathcal{G}_{\theta_r}\!\Big(\mathcal{E}_{\theta_v}(s^n, f^n)\Big)$,
    \State Sample $\tau \sim \mathrm{U}(1,\mathrm{T})$, $\epsilon \sim \mathcal{N}(0,\mathbf{I})$,
    \State Sample $\boldsymbol{z}^{n}_{\tau}$ at $\tau$ by
    $\boldsymbol{z}^{n}_{\tau} \gets q(\boldsymbol{z}^{n}_{\tau} \mid \boldsymbol{z}^{n}_0)$ via Eq.~\eqref{eq:dif-forward-onestep-det},
    \State Optimize ${\mathbf{Z}_\theta}$ by taking a gradient descent step on Eq.~\eqref{eq:IGG}.
\Until{converged}
\end{algorithmic}
\end{algorithm}

\begin{algorithm}[!h]
\caption{IGG Sampling}
\label{alg:sampling}
\begin{algorithmic}[1]
\Require Data $\{ s^n, \Lambda^n \}_{n=1}^{N_{\text{test}}}$
\Ensure Geodesic deformations $\boldsymbol{\Phi}^n$

\State Initialize $\boldsymbol{z}^{n}_{T} \sim \mathcal{N}(0,\mathbf{I})$
\For{$\tau = \mathrm{T}, \ldots, 1$}
    \State $\hat{\boldsymbol{z}}^{n}_{\tau-1} \gets
    p_{\theta}\!\Big(\boldsymbol{z}^{n}_{\tau-1} \mid \boldsymbol{z}^{n}_{\tau}, \nabla s^{n}, \Lambda^{n}, \tau\Big)$
    via Eq.~\eqref{eq:dif-back},
\EndFor
\State $\{\hat{\boldsymbol{v}}^n\} \gets \mathcal{D}_{\theta_v}(\hat{\boldsymbol{z}}^{n}_{0})$,
\State $\boldsymbol{\Phi}^n$ $\gets \{\hat{\boldsymbol{v}}^n\}$ via Eq.~\eqref{eq:phi_v},
\State \Return $\boldsymbol{\Phi}^n$
\end{algorithmic}
\end{algorithm}

\section{Experimental Evaluation}
We evaluate the proposed IGG framework using a diverse set of real-world image datasets that capture deformable shape changes over time. We first assess the quality of the learned latent representations of geodesics by comparing them to numerical solutions of the EPDiff equation obtained via Euler integration. 

Next, we evaluate the quality of the generated images deformed by the sampled geodesics guided by given text instructions. These results are compared with state-of-the-art generative models that synthesize image sequences/videos with publicly available training code or fine-tuning options, including the video diffusion model (VDM)~\citep{ho2022video}, CogVideoX~\citep{yang2024cogvideox}, and DynamiCrafter~\citep{xing2025dynamicrafter}. Note that all baselines are trained using sequences of deformed images generated along the geodesic transformations derived from the numerical solutions of EPDiff equation. This approach ensures that the results are independent of the learned geodesics predicted by the first module of the IGG framework, providing a consistent and unbiased evaluation.

Finally, to further assess the quality of the generated samples and demonstrate their clinical utility in real-world applications, we design a downstream task in the context of brain MRI segmentation. More specifically, we use the synthesized image samples as augmentation data by pairing each transformed image with its corresponding segmentation label, both generated through the learned deformation fields applied to existing images and labels. We conduct experiments using a diverse set of widely used segmentation architectures, including U-Net~\citep{ronneberger2015u}, TransUNet~\citep{chen2021transunet}, and Attention U-Net (AttUNet)~\citep{oktay2018attention}, under limited training data conditions.

\subsection{Dataset}
\noindent{\bf Komatsuna plant.} We include $300$ frames of RGB-D label-maps representing five Komatsuna plants from the publicly available data repository~\citep{uchiyama2017easy}. The label maps cover different leaves that emerge from the bud and grow in size over time, where the plant growth was monitored between $228$-$236$ hours. All data frames were resampled to to $128^2$ and pre-aligned with affine transformations. \\

\noindent{\bf Longitudinal brain MRI.} We include a total of $2618$ T1-weighted longitudinal (time-series) brain MRIs sourced from the Open Access Series of Imaging Studies (OASIS-3) dataset~\citep{lamontagne2019oasis}. This experiment aims to validate our method using longitudinal data that includes scans at varying time intervals for individuals spanning both healthy subjects and Alzheimer's diseases (AD), aged $60$-$90$. Given the scenario that many existing image generation/editing methods with text instructions focus on 2D natural images~\citep{brooks2023instructpix2pix,meng2021sdedit}, we specifically utilize 2D scans derived from this 3D brain data for comparison with state-of-the-art baselines. All MRIs undergo pre-processing, including resizing to $128^2$, with isotropic voxels of $1 \text{mm}^2$, skull-stripping, intensity normalization, bias field correction, and pre-alignment using affine transformations. \\

\noindent{\bf Text instructions.} Our text condition consists of a description of the template image, followed by details of the specific edits or progressions applied to it. For brain MRIs, this includes biological variables such as age, sex, and gender. In contrast, for plants, it primarily focuses on the growth timeline. All information is sourced directly from the original dataset available in the data repository.

\subsection{Experimental Design and Implementation Details}
\noindent{\bf Evaluate learned latent representations of geodesics via GIR.} We evaluate GIR's ability to predict geodesic dynamics by comparing the learned geodesics with numerical solutions to the EPDiff equation using Euler integration. The assessment focuses on errors in velocity fields, transformations, and deformed images between the two approaches. Quantitative metrics, such as the mean absolute error of velocity fields, are computed at each time integration step. \\

\noindent{\bf Evaluate generated samples.} We compare the quality of samples generated by IGG with three baseline models: VDM~\citep{ho2022video}, CogVideoX~\citep{yang2024cogvideox}, and DynamiCrafter~\citep{xing2025dynamicrafter}). In particular, we first assess individual synthesized images at each time step using standard image synthesis metrics, including the Fréchet Inception Distance (FID)~\citep{heusel2017gans} and Kernel Inception Distance (KID)~\citep{binkowski2018demystifying} to measure the distributional similarity between generated and real images. Additionally, we use the Structural Similarity Index (SSIM)~\citep{wang2004image} to evaluate structural, contrast, and luminance similarities. We then evaluate the continuity and smoothness of generated video sequences by computing the Fréchet Video Distance (FVD)~\citep{unterthiner2018towards} and providing qualitative visual comparisons. Finally, we assess the perceptual quality of the generated samples using the Inception Score (IS)~\citep{salimans2016improved}. \\

\noindent{\bf Evaluate the preservation of object geometry and topology.} To demonstrate the efficiency of our IGG model in preserving fine-grained geometric structure and topology, we first generate image sequences by deforming a template image using the sampled geodesic flow of deformations produced by IGG and compare these results with all baseline models. Secondly, we present the corresponding determinant of the Jacobian (DetJac) maps for the deformations. The DetJac values reveal important patterns of volume change: a value of $1$ indicates no volume change, DetJac $<1$ reflects volume shrinkage, and DetJac $>1$ implies volume expansion. A DetJac value below zero indicates artifacts or singularities in the transformation field, highlighting a failure to maintain the topological integrity. Note that only our IGG model enables such a metric to quantify the topological changes in the generated samples. \\

\begin{figure*}[!b]
\centering
\includegraphics[width=1.0\textwidth] {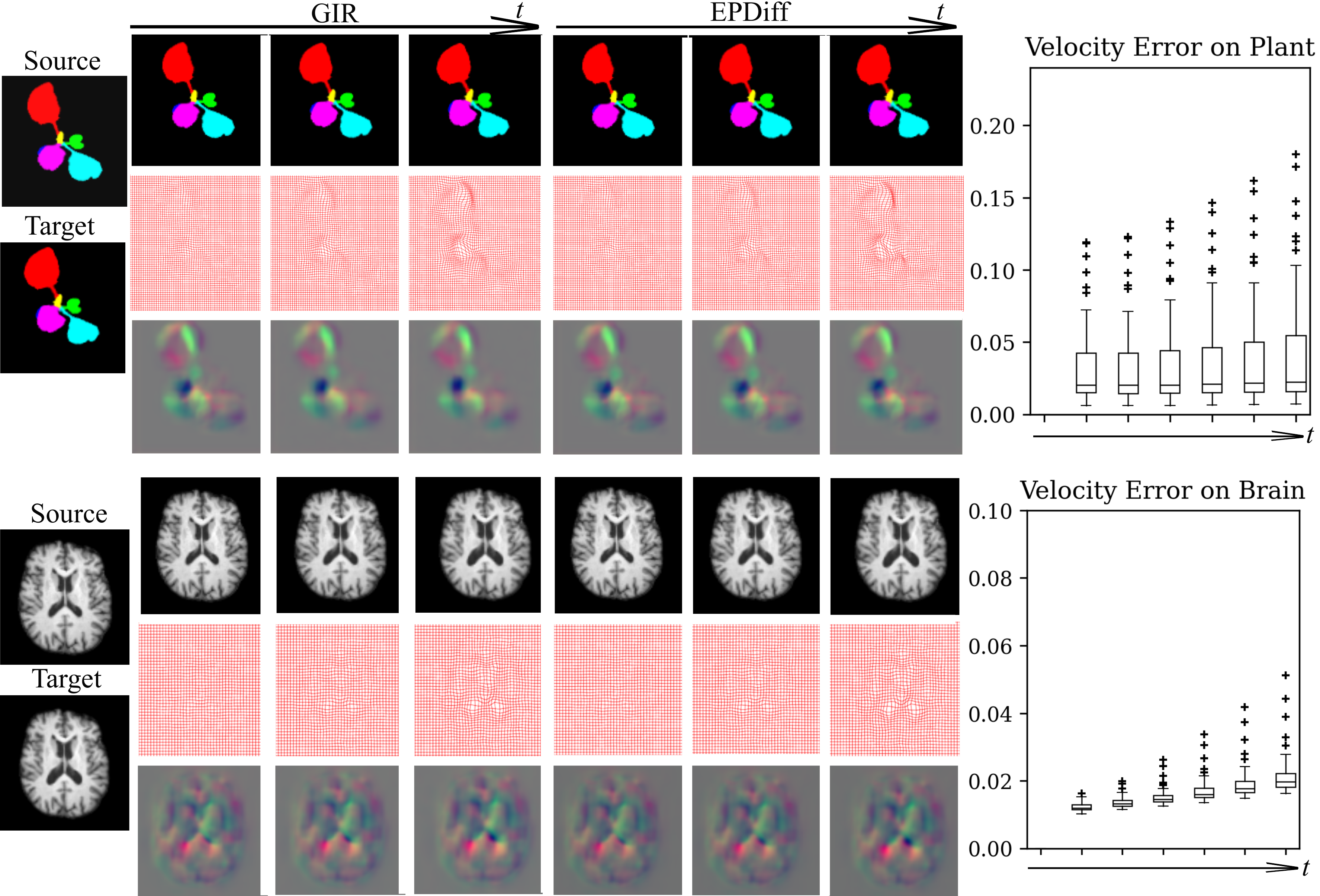}
     \caption{Comparison of predicted geodesics by GIR (the first sub-module of IGG) vs. real numerical solutions from the EPDiff equation. Left: Visualization of predicted deformed images, deformations, and velocities along time. Right: Mean absolute error of predicted velocities over time compared to numerical integration of EPDiff.}
\label{fig:CompareAE}
\end{figure*}

\noindent{\bf Evaluate reliability of model predictions.} We evaluate the reliability and confidence of IGG in learning the geodesic deformations of geometric shapes over time by computing pixel-wise mean and standard deviation for individual sampled image sequences along the geodesic path. We visualize confidence intervals to highlight regions with 95\% certainty in growth patterns. These intervals are defined by the mean of $1000$ generated samples with pixel-wise bounds that are two standard deviations above and below the mean. \\

\noindent{\bf Demonstrate utility of generated samples in downstream segmentation tasks.} We evaluate the effectiveness of IGG generated images by using them as augmented data for segmentation tasks under conditions of limited number of training samples. We then compare the performance of segmentation models trained with and without our synthesized augmented data, considering dataset augmentations of $2\times$ and $4\times$ the size of the original dataset. The segmentation accuracy is assessed using the dice score~\citep{dice1945measures}, both averaged across all anatomical regions and separately for different structures. To highlight the flexibility of IGG in generating new data, we evaluate augmentation performance on OASIS-3~\citep{lamontagne2019oasis} and OASIS-1~\citep{marcus2007open} datasets independently, noting that our model is trained exclusively on OASIS-3. This setup allows us to examine the generalization capability of the synthesized data to an unseen dataset while quantifying its impact on segmentation performance. \\

\noindent{\bf Parameter Setting.} We split all datasets into $80\%$, $10\%$, and $10\%$ for training, validation, and testing. All experiments are conducted on NVIDIA A100 GPUs. For training the autoencoder for geodesic learning, we alternatively update the encoder-decoder and the latent geodesic neural operator, and finally jointly train both components for $2000$ epochs per stage. We set the weight decay as $1e^{-4}$, batch size as $64$, and a learning rate as $5e^{-4}$. For training the latent geodesic diffusion model, we set the learning rate as $1e^{-4}$, the number of epochs as $6000$, batch size as $36$, and diffusion steps as $500$. All models are trained using the Adam optimizer~\citep{kingma2014adam}.

\subsection{Experimental Results}

The left panel of Fig.~\ref{fig:CompareAE} visualizes the predicted transformations, velocity fields, and deformed images from IGG's geodesic-informed registration network (GIR), compared with numerical solutions to the EPDiff equation (Eq.~\eqref{eq:epdiff}). These methods show a high degree of similarity. The right panel of Fig.~\ref{fig:CompareAE} presents quantitative results of the absolute per-pixel error, computed between the velocity field obtained from the numerical solution of the EPDiff equation and the velocity field predicted by the GIR module at each time step. These results collectively highlight GIR's ability to effectively learn geodesic mapping functions comparable to real numerical solutions.

\begin{figure*}[!t]
\centering
\includegraphics[width=0.85\textwidth] {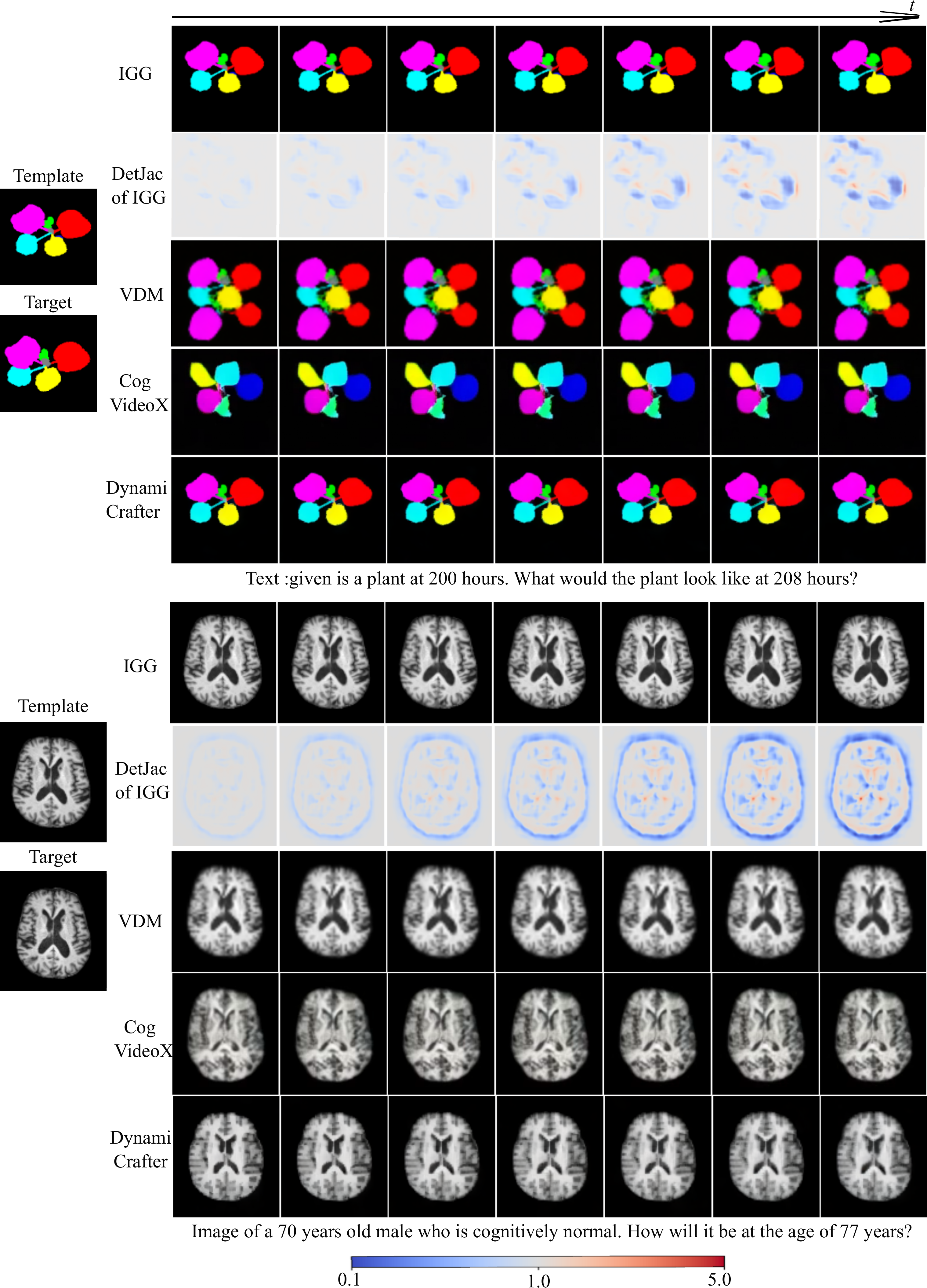}
     \caption{A comparison of images generated by IGG (with corresponding DetJac) against all baseline models across different time frames. Given an input template image and text instructions, all models generate samples of target images. The ground truth "target" along with input template images are provided on the left side of the panel for reference.}
\label{fig:CompareGeo}
\end{figure*}

\begin{table*}[!b]
\centering
\caption{A comparison of performance metrics across all methods for various datasets.}
\resizebox{0.9\textwidth}{!}{%
\begin{tabular}{lcccccc}
\toprule
Dataset & Model & FVD $\downarrow$ & FID $\downarrow$ & KID $\downarrow$ & SSIM $\uparrow$ & IS $\uparrow$ \\
\midrule
\multirow{4}{*}{Plant} 
    & IGG & $\textbf{48.29}$ & $\textbf{9.23}$ & $\textbf{0.007} \pm \textbf{0.02}$ & $\textbf{0.98} \pm \textbf{0.02}$ & $\textbf{1.06} \pm \textbf{1.2}\%$ \\
    & VDM~\citep{ho2022video} & $357.56$ & $71.79$ & $0.23 \pm 0.03$ & $0.23 \pm 0.10$ & $1.05 \pm 1.0\%$ \\
    & CogVideoX~\citep{yang2024cogvideox} & $471.49$ & $153.93$ & $0.63 \pm 0.07$ & $0.69 \pm 0.19$ & $1.05 \pm 0.6\%$ \\
    & DynamiCrafter~\citep{xing2025dynamicrafter} & $466.70$ & $80.78$ & $0.41 \pm 0.04$ & $0.89 \pm 0.04$ & $1.03 \pm 0.3\%$ \\
\midrule
\multirow{4}{*}{Brain} 
    & IGG & $\textbf{26.39}$ & $\textbf{6.23}$ & $\textbf{0.006} \pm \textbf{0.008}$ & $\textbf{0.97} \pm \textbf{0.03}$ & $1.02 \pm 0.08\%$ \\
    & VDM~\citep{ho2022video} & $247.84$ & $69.23$ & $0.46 \pm 0.03$ & $0.89 \pm 0.02$ & $1.02 \pm 0.07\%$ \\
    & CogVideoX~\citep{yang2024cogvideox} & $302.20$ & $114.50$ & $0.46 \pm 0.04$ & $0.69 \pm 0.27$ & $\textbf{1.06} \pm \textbf{2.8}\%$ \\
    & DynamiCrafter~\citep{xing2025dynamicrafter} & $288.62$ & $142.40$ & $1.05 \pm 0.03$ & $0.88 \pm 0.03$ & $1.03 \pm 0.20\%$ \\
\bottomrule
\end{tabular}%
}
\label{table_comp}
\end{table*}

We also report the additional computational overhead introduced by the GIR, which primarily comes from the latent neural operator used to approximate the EPDiff equation (Eq.~\eqref{eq:epdiff}). In practice, in our experiments on $128^2$ brain images, the training time per epoch is approximately $13.10$s for GIR, compared to $4.81$s for a standard registration model with numerical solutions to the EPDiff~\cite{hinkle2018diffeomorphic}. The corresponding inference times are $0.01$s vs. $0.004$s, respectively.

Fig.~\ref{fig:CompareGeo} compares the generated images from our model IGG (along with the associated DetJac maps of the deformations) with the baseline generative models. The visualized images and DetJac values of IGG suggest that our model effectively preserves the topological structure of objects within the generated images with well-captured progression of geometric shape changes over time. In contrast, samples generated by the baselines fail to maintain the geometric integrity of various structures. For instance, in plant growth images, the leaves appear to merge or overlap unnaturally, disrupting their biological topology. Similarly, the baselines inaccurately predict parts of the ventricles, resulting in regions that do not correspond to the original brain structure. It is worth noting that the baseline methods perform diffusion directly in the image intensity space, learning the distribution of raw pixel values. During sampling, they generate pixel intensities through iterative denoising, which can lead to blurring (or noisy) artifacts observed in diffusion models operating in pixel space. In contrast, our method performs diffusion in the deformation space, generating velocity or deformation fields that are subsequently applied to a template image to synthesize new image sequences. By modeling structural variation through geometric transformations rather than pixel-wise generation, our approach better preserves sharp and anatomically coherent structures.

Tab.~\ref{table_comp} reports the evaluation metrics of sample quality and diversity for the generated images from our model IGG and the baselines. While all models achieve similar IS, IGG shows significantly lower FVD, FID, and KID scores, and higher SSIM scores. This indicates that while all models generate diverse images (leading to a good IS), the baselines lack realistic features or proper distribution alignment with real images. The observation of IGG achieving approximately $10$ times better scores demonstrates its ability to generate more realistic images.

\begin{figure*}[!h]
\centering
\includegraphics[width=1.0\textwidth] {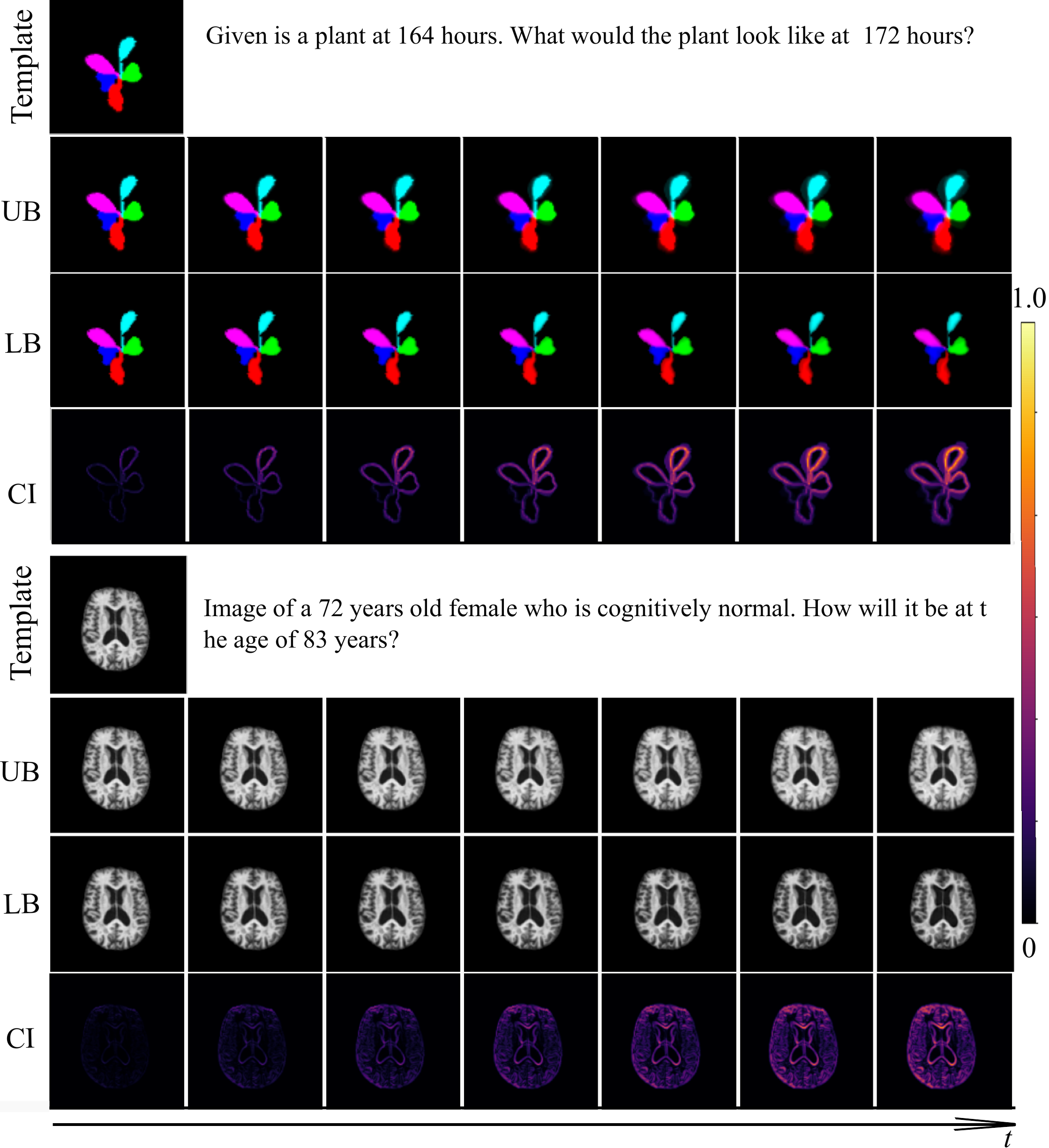}
     \caption{Top: Input template images with text instructions. Bottom: Confidence maps illustrating the lower bounds (LB), upper bounds (UB), and confidence intervals (CI), which represent regions representing 95\% of ideal growth patterns, based on ~1000 samples generated by our IGG model across different time frames.}
\label{fig:CDM-PANT-OASIS}
\end{figure*}

Fig.~\ref{fig:CDM-PANT-OASIS} visualizes examples of 
confidence maps of plant growth and brain progression images generated from IGG. It suggests that our model effectively captures the growth pattern of plants over time, primarily focusing on the boundaries of the leaves. The confidence maps of brain images highlight expanding patterns in ventricles, demonstrating the model's ability to capture dynamic changes over time.

\begin{figure*}[!h]
\centering
\includegraphics[width=0.95\textwidth] {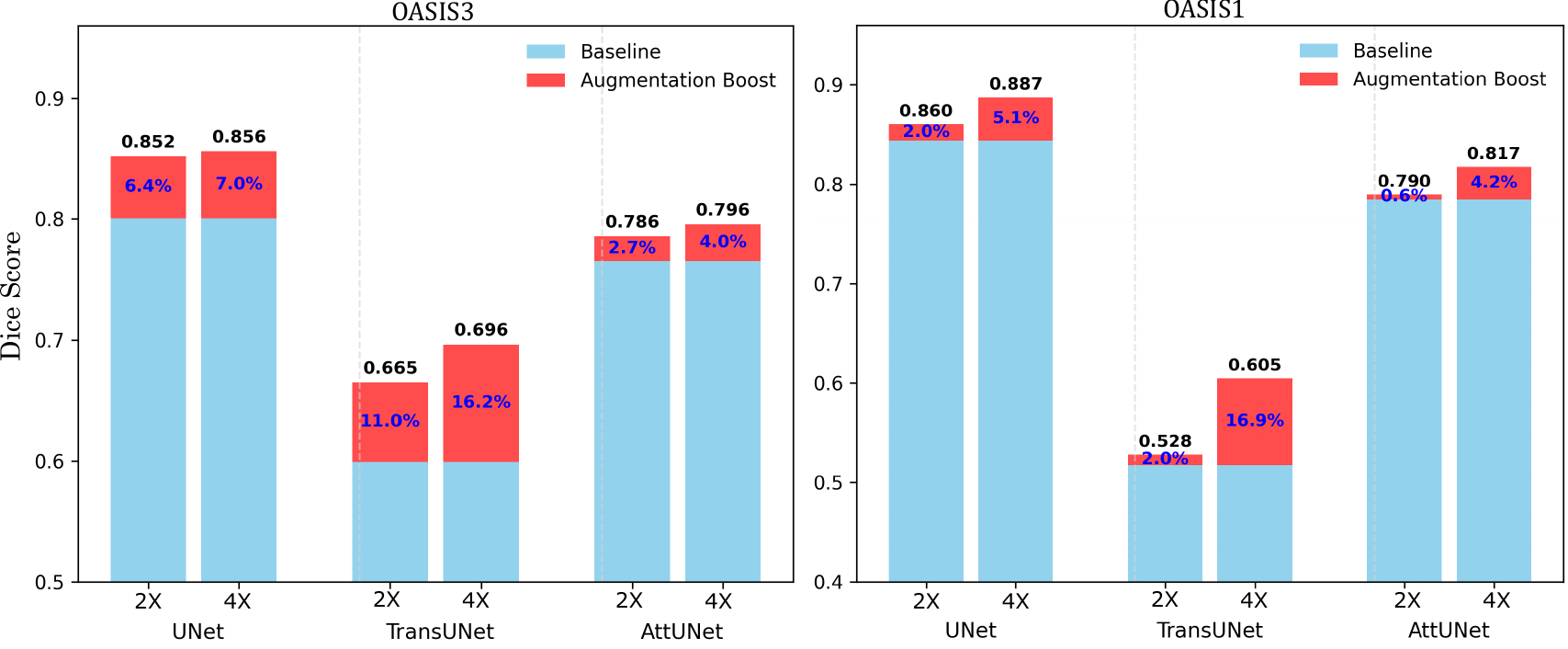}
     \caption{The improvement in average dice scores achieved by different segmentation models when trained with varying training dataset configurations. Baseline denotes training with the original limited dataset, while $2\times$ and $4\times$ correspond to augmentations of $2$ times and $4$ times the size of the original dataset, respectively.}
\label{fig:2D-dice-Imp}
\end{figure*}

\begin{figure*}[!h]
\centering
\includegraphics[width=0.95\textwidth] {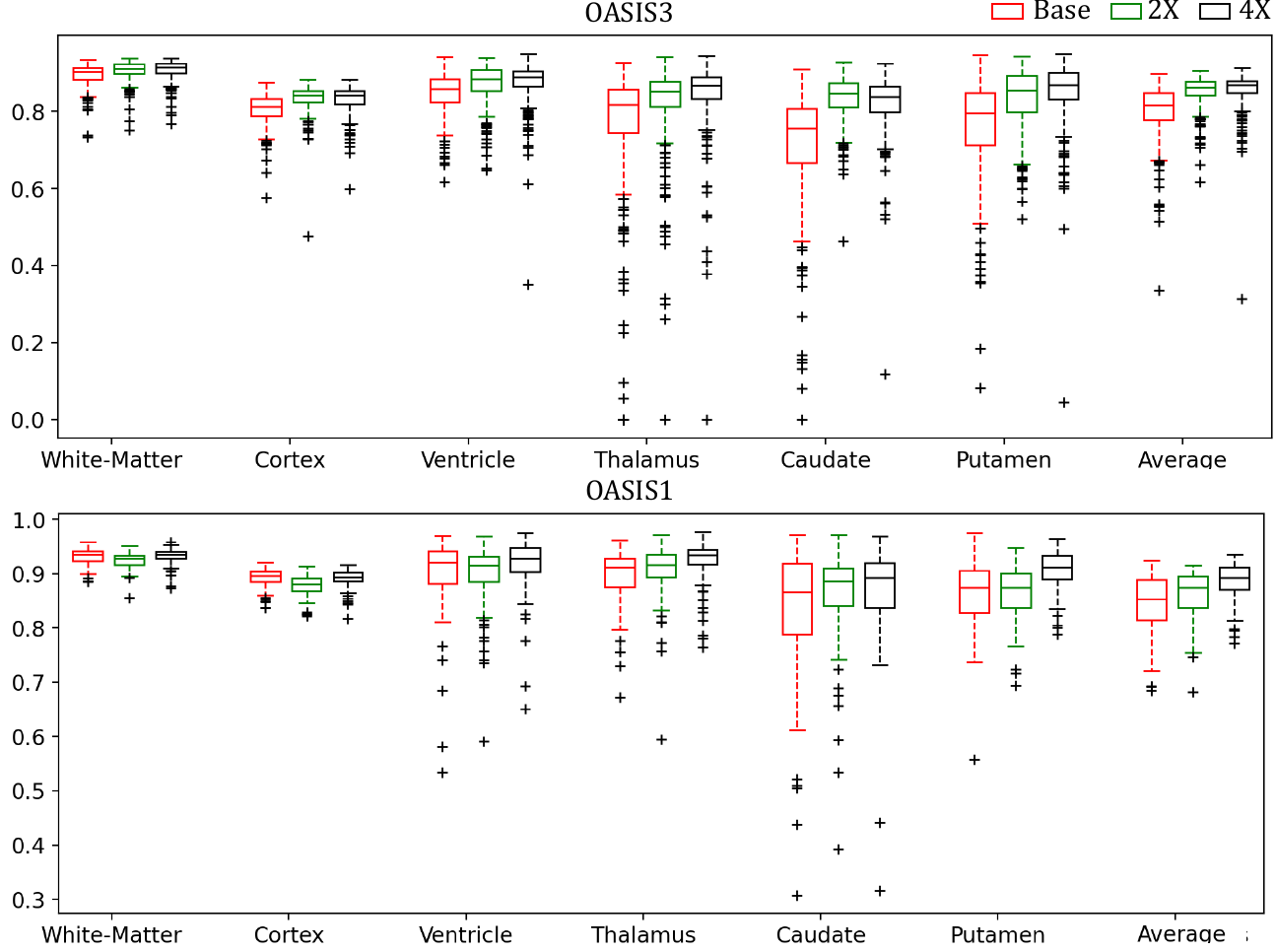}
     \caption{A comparison of average dice scores across different anatomical structures under varying training dataset configurations. Left to right: {\bf base:} training with the original limited dataset, and $2\times$ and $4\times$ correspond to augmented sizes of the original dataset. Top to bottom: results on the OASIS-3 vs. OASIS-1.}
\label{fig:2D-dice-seg}
\end{figure*}

Fig.~\ref{fig:2D-dice-Imp} presents the improvement in average dice scores for different segmentation models trained with varying dataset configurations. The baseline corresponds to training on the original limited dataset. It shows that with gradually increasing the amount of augmented data generated by IGG from $2 \times$ to $4 \times$ the size of the original training set, we can consistently obtain better segmentation performance with higher dice scores across all models. These results demonstrate that the generated image-label pairs are able to effectively augment real data, particularly under limited-label conditions. While the improvements on OASIS-1 (with maximal $+5.1$\%) are smaller than those on OASIS-3 (with maximal $+11.0$\%), they remain consistent, highlighting the generalizability of our model across datasets.

Fig.~\ref{fig:2D-dice-seg} further reports detailed dice scores across different anatomical structures when using IGG’s predictions as augmented data on both OASIS-3 and OASIS-1 datasets.

\begin{figure*}[!h]
\centering
\includegraphics[width=1.0\textwidth] {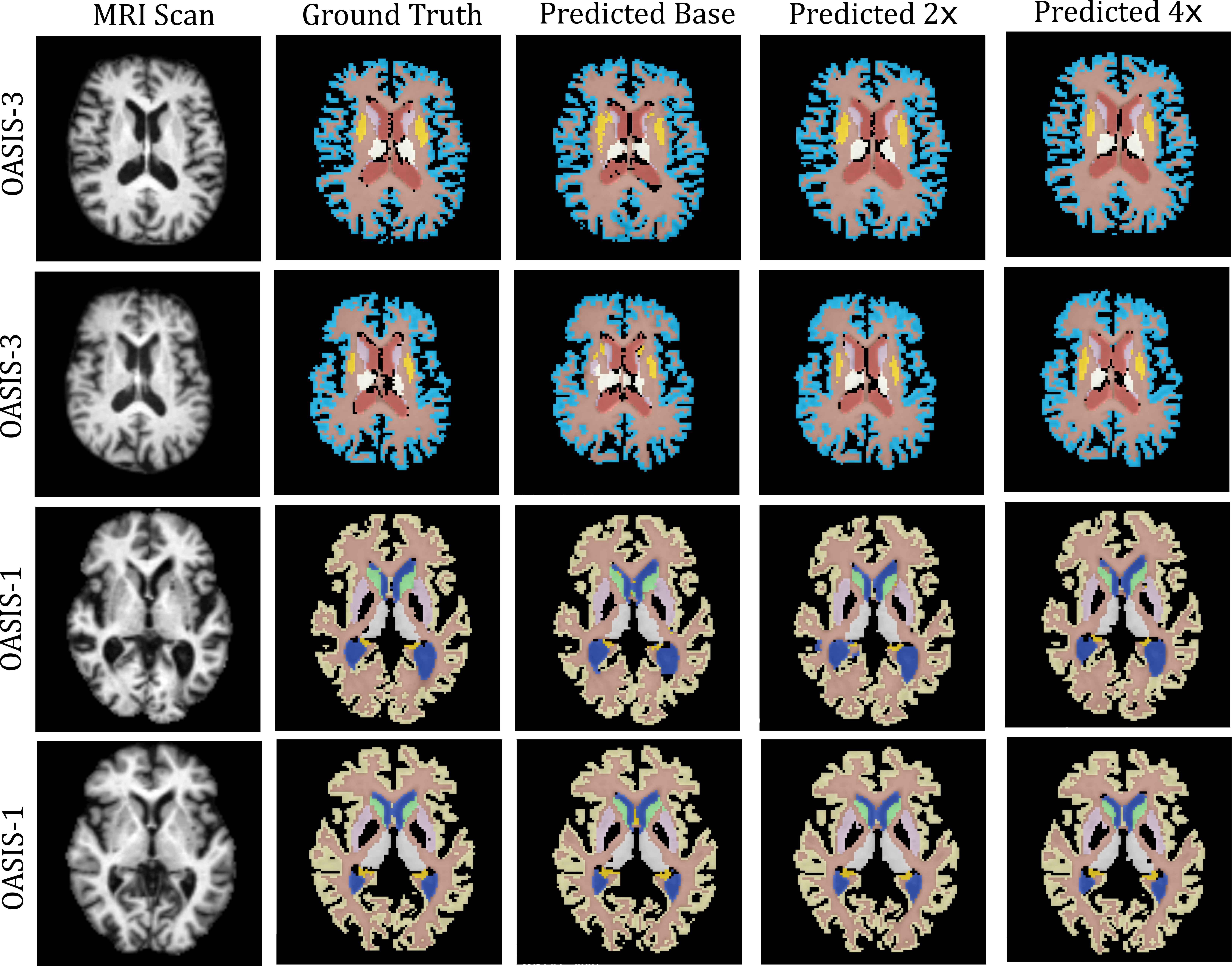}
     \caption{Examples of segmentation results on brain MRIs across different anatomical structures under varying training dataset configurations. Left to right: original MRI scans, ground truth labels, predicted labels from a model trained on the original limited dataset, and predicted labels from models trained with $2\times$ and $4\times$ the size of the original dataset. 
     Top to bottom: results on the OASIS-3 vs. OASIS-1.}
\label{fig:labels_seg}
\end{figure*}

Fig.~\ref{fig:labels_seg} visualizes representative segmentation maps predicted by a trained U-Net~\citep{ronneberger2015u} in different configurations of the training dataset. When trained on a limited dataset, and the model produces errors in certain anatomical structures, whereas training with augmented data yields predictions that more closely match the ground truth. These results further confirm the effectiveness of incorporating our generated samples, especially under limited-data conditions. 


\section{Conclusion \& Discussion}
This paper presents an extended work, IGG~\citep{wu2025igg}, a geodesic-informed generative diffusion model for topology-preserved image video generation. In contrast to existing generative models that focus primarily on manipulating image intensity and texture, IGG explicitly learns the underlying latent distribution of geodesic-informed deformations from images, enabling the synthesis of diverse samples within geodesic deformation spaces. Our work is the first generative model to operate on natural and interpretable image transformations: it not only captures shape-changing processes during generation, but also provides quantitative metrics to assess topological consistency in the synthesized samples. The key contributions of IGG include (i)
a geodesic-informed registration network that learns latent representations of time-dependent diffeomorphic transformations, enabling accurate modeling of complex shape dynamics; and (ii) a latent geometric diffusion model that captures sequential deformation features conditioned on text inputs and template images, allowing controlled and semantically meaningful generation. Our experimental results demonstrate that IGG significantly improves sample fidelity and diversity compared to state-of-the-art methods. Additionally, downstream segmentation experiments show substantial performance gains when using IGG-generated samples as data augmentation, particularly in low-data regimes.

Our proposed IGG model advances image generation by integrating geodesic principles, providing tools for topology assessment, and establishing a framework for synthesizing anatomically consistent image samples. We acknowledge that the fidelity of the synthesized data from our IGG model is inherently dependent on the accuracy of the underlying registration. When registration errors are large, the quality of the synthesized results may be correspondingly limited. This dependency reflects a critical assumption of our framework that anatomically meaningful correspondences can be reliably established. Our topological prior is most effective in applications where anatomical structures are expected to preserve diffeomorphic consistency. However, in practical settings, imaging noise, pathology-induced appearance changes, or other non-diffeomorphic structural changes may violate this assumption. In such cases, strict topology preservation may not fully capture the underlying anatomical variability. Our future work will be investigated to combine IGG’s strength in learning deformation fields with the modeling of image texture distributions, which will enable the development of more powerful and realistic generative models that capture both structural and appearance variations. Other future research will focus on several potential directions in extending IGG to handle 4D data (3D + time) to model dynamic longitudinal anatomical changes, such as cardiac motion or brain changes over time. These directions aim to further enhance the clinical and scientific utility of generative modeling in medical imaging.

\acks{This work was supported by NSF CAREER Grant 2239977.}

\ethics{The work follows appropriate ethical standards in conducting research and writing the manuscript, following all applicable laws and regulations regarding treatment of animals or human subjects.}

\coi{We declare we do not have conflicts of interest.}

\data{The Brain MRIs used in training and testing in this paper are from the Open Access Series of Imaging Studies (OASIS-3) dataset~\citep{lamontagne2019oasis}, which are readily accessible and user-friendly. 
Readers interested in evaluating the accuracy of the method can access and utilize the publicly available datasets. The code for this study is also publicly available at \url{https://github.com/nellie689/IGG}.}

\bibliography{main}

\begin{thebibliography}{69}
\providecommand{\natexlab}[1]{#1}
\providecommand{\url}[1]{\texttt{#1}}
\expandafter\ifx\csname urlstyle\endcsname\relax
  \providecommand{\doi}[1]{doi: #1}\else
  \providecommand{\doi}{doi: \begingroup \urlstyle{rm}\Url}\fi

\bibitem[Arnold(1966)]{arnold1966}
Vladimir Arnold.
\newblock Sur la g{\'e}om{\'e}trie diff{\'e}rentielle des groupes de lie de dimension infinie et ses applications {\`a} l'hydrodynamique des fluides parfaits.
\newblock In \emph{Annales de l'institut Fourier}, volume~16, pages 319--361, 1966.

\bibitem[Avants et~al.(2008)Avants, Epstein, Grossman, and Gee]{avants2008symmetric}
Brian~B Avants, Charles~L Epstein, Murray Grossman, and James~C Gee.
\newblock Symmetric diffeomorphic image registration with cross-correlation: evaluating automated labeling of elderly and neurodegenerative brain.
\newblock \emph{Medical image analysis}, 12\penalty0 (1):\penalty0 26--41, 2008.

\bibitem[Azizi et~al.(2023)Azizi, Kornblith, Saharia, Norouzi, and Fleet]{azizi2023synthetic}
Shekoofeh Azizi, Simon Kornblith, Chitwan Saharia, Mohammad Norouzi, and David~J Fleet.
\newblock Synthetic data from diffusion models improves imagenet classification.
\newblock \emph{arXiv:2304.08466}, 2023.

\bibitem[Bang and Shim(2021)]{bang2021mggan}
Duhyeon Bang and Hyunjung Shim.
\newblock Mggan: Solving mode collapse using manifold-guided training.
\newblock In \emph{Proceedings of the IEEE/CVF international conference on computer vision}, pages 2347--2356, 2021.

\bibitem[Beg et~al.(2005)Beg, Miller, Trouv{\'e}, and Younes]{beg2005computing}
Mirza~Faisal Beg, Michael~I Miller, Alain Trouv{\'e}, and Laurent Younes.
\newblock Computing large deformation metric mappings via geodesic flows of diffeomorphisms.
\newblock \emph{International journal of computer vision}, 61\penalty0 (2):\penalty0 139--157, 2005.

\bibitem[Bi{\'n}kowski et~al.(2018)Bi{\'n}kowski, Sutherland, Arbel, and Gretton]{binkowski2018demystifying}
Miko{\l}aj Bi{\'n}kowski, Danica~J Sutherland, Michael Arbel, and Arthur Gretton.
\newblock Demystifying mmd gans.
\newblock \emph{arXiv:1801.01401}, 2018.

\bibitem[Brooks et~al.(2023)Brooks, Holynski, and Efros]{brooks2023instructpix2pix}
Tim Brooks, Aleksander Holynski, and Alexei~A Efros.
\newblock Instructpix2pix: Learning to follow image editing instructions.
\newblock In \emph{Proceedings of the IEEE/CVF Conference on Computer Vision and Pattern Recognition}, pages 18392--18402, 2023.

\bibitem[Chen et~al.(2021)Chen, Lu, Yu, Luo, Adeli, Wang, Lu, Yuille, and Zhou]{chen2021transunet}
Jieneng Chen, Yongyi Lu, Qihang Yu, Xiangde Luo, Ehsan Adeli, Yan Wang, Le~Lu, Alan~L Yuille, and Yuyin Zhou.
\newblock Transunet: Transformers make strong encoders for medical image segmentation.
\newblock \emph{arXiv preprint arXiv:2102.04306}, 2021.

\bibitem[Dao et~al.(2024)Dao, Yang, and Kim]{dao2024conditional}
Duy-Phuong Dao, Hyung-Jeong Yang, and Jahae Kim.
\newblock Conditional diffusion model for longitudinal medical image generation.
\newblock \emph{arXiv preprint arXiv:2411.05860}, 2024.

\bibitem[Deb et~al.(2025)Deb, Wu, Epstein, and Zhang]{deb2025unsupervised}
Swakshar Deb, Nian Wu, Frederick~H Epstein, and Miaomiao Zhang.
\newblock Unsupervised cardiac video translation via motion feature guided diffusion model.
\newblock In \emph{International Conference on Medical Image Computing and Computer-Assisted Intervention}, pages 648--658. Springer, 2025.

\bibitem[Dice(1945)]{dice1945measures}
Lee~R Dice.
\newblock Measures of the amount of ecologic association between species.
\newblock \emph{Ecology}, 26\penalty0 (3):\penalty0 297--302, 1945.

\bibitem[Goodfellow et~al.(2020)Goodfellow, Pouget-Abadie, Mirza, Xu, Warde-Farley, Ozair, Courville, and Bengio]{goodfellow2020generative}
Ian Goodfellow, Jean Pouget-Abadie, Mehdi Mirza, Bing Xu, David Warde-Farley, Sherjil Ozair, Aaron Courville, and Yoshua Bengio.
\newblock Generative adversarial networks.
\newblock \emph{Communications of the ACM}, 63\penalty0 (11):\penalty0 139--144, 2020.

\bibitem[Graf et~al.(2023)Graf, Schmitt, Schlaeger, M{\"o}ller, Sideri-Lampretsa, Sekuboyina, Krieg, Wiestler, Menze, Rueckert, et~al.]{graf2023denoising}
Robert Graf, Joachim Schmitt, Sarah Schlaeger, Hendrik~Kristian M{\"o}ller, Vasiliki Sideri-Lampretsa, Anjany Sekuboyina, Sandro~Manuel Krieg, Benedikt Wiestler, Bjoern Menze, Daniel Rueckert, et~al.
\newblock Denoising diffusion-based mri to ct image translation enables automated spinal segmentation.
\newblock \emph{European Radiology Experimental}, 7\penalty0 (1):\penalty0 70, 2023.

\bibitem[Gupta et~al.(2024)Gupta, Samaras, and Chen]{gupta2024topodiffusionnet}
Saumya Gupta, Dimitris Samaras, and Chao Chen.
\newblock Topodiffusionnet: A topology-aware diffusion model.
\newblock \emph{arXiv:2410.16646}, 2024.

\bibitem[Hendrycks and Gimpel(2016)]{hendrycks2016gaussian}
Dan Hendrycks and Kevin Gimpel.
\newblock Gaussian error linear units (gelus).
\newblock \emph{arXiv preprint arXiv:1606.08415}, 2016.

\bibitem[Hertz et~al.(2022)Hertz, Mokady, Tenenbaum, Aberman, Pritch, and Cohen-Or]{hertz2022prompttoprompt}
Amir Hertz, Ron Mokady, Jay Tenenbaum, Kfir Aberman, Yael Pritch, and Daniel Cohen-Or.
\newblock Prompt-to-prompt image editing with cross attention control, 2022.

\bibitem[Heusel et~al.(2017)Heusel, Ramsauer, Unterthiner, Nessler, and Hochreiter]{heusel2017gans}
Martin Heusel, Hubert Ramsauer, Thomas Unterthiner, Bernhard Nessler, and Sepp Hochreiter.
\newblock Gans trained by a two time-scale update rule converge to a local nash equilibrium.
\newblock \emph{Advances in neural information processing systems}, 30, 2017.

\bibitem[Hinkle et~al.(2018)Hinkle, Womble, and Yoon]{hinkle2018diffeomorphic}
J.~Hinkle, D.~Womble, and H.J.d Yoon.
\newblock Diffeomorphic autoencoders for {LDDMM} atlas building.
\newblock In \emph{Medical Imaging with Deep Learning}, 2018.

\bibitem[Ho and Salimans(2022)]{ho2022classifier}
Jonathan Ho and Tim Salimans.
\newblock Classifier-free diffusion guidance.
\newblock \emph{arXiv:2207.12598}, 2022.

\bibitem[Ho et~al.(2020)Ho, Jain, and Abbeel]{ho2020denoising}
Jonathan Ho, Ajay Jain, and Pieter Abbeel.
\newblock Denoising diffusion probabilistic models.
\newblock \emph{Advances in Neural Information Processing Systems}, 33:\penalty0 6840--6851, 2020.

\bibitem[Ho et~al.(2022)Ho, Salimans, Gritsenko, Chan, Norouzi, and Fleet]{ho2022video}
Jonathan Ho, Tim Salimans, Alexey Gritsenko, William Chan, Mohammad Norouzi, and David~J Fleet.
\newblock Video diffusion models.
\newblock \emph{Advances in Neural Information Processing Systems}, 35:\penalty0 8633--8646, 2022.

\bibitem[Hossain and Zhang(2025)]{hossain2025mgaug}
Tonmoy Hossain and Miaomiao Zhang.
\newblock Mgaug: Multimodal geometric augmentation in latent spaces of image deformations.
\newblock \emph{Medical Image Analysis}, 102:\penalty0 103540, 2025.

\bibitem[Hu et~al.(2024)Hu, Fei, Xu, Hou, Yang, Wang, Lei, Qian, and He]{hu2024topology}
Jiangbei Hu, Ben Fei, Baixin Xu, Fei Hou, Weidong Yang, Shengfa Wang, Na~Lei, Chen Qian, and Ying He.
\newblock Topology-aware latent diffusion for 3d shape generation.
\newblock \emph{arXiv:2401.17603}, 2024.

\bibitem[Jayakumar et~al.(2023)Jayakumar, Hossain, and Zhang]{jayakumar2023sadir}
Nivetha Jayakumar, Tonmoy Hossain, and Miaomiao Zhang.
\newblock Sadir: shape-aware diffusion models for 3d image reconstruction.
\newblock In \emph{International workshop on shape in medical imaging}, pages 287--300. Springer, 2023.

\bibitem[Jayakumar et~al.(2024)Jayakumar, Gadila, Hossain, Ji, and Zhang]{jayakumar2024tpie}
Nivetha Jayakumar, Srivardhan~Reddy Gadila, Tonmoy Hossain, Yangfeng Ji, and Miaomiao Zhang.
\newblock Tpie: Topology-preserved image editing with text instructions.
\newblock \emph{arXiv preprint arXiv:2411.16714}, 2024.

\bibitem[Joshi et~al.(2004)Joshi, Davis, Jomier, and Gerig]{joshi2004unbiased}
Sarang Joshi, Brad Davis, Matthieu Jomier, and Guido Gerig.
\newblock Unbiased diffeomorphic atlas construction for computational anatomy.
\newblock \emph{NeuroImage}, 23:\penalty0 S151--S160, 2004.

\bibitem[Kawar et~al.(2023)Kawar, Zada, Lang, Tov, Chang, Dekel, Mosseri, and Irani]{kawar2023imagic}
Bahjat Kawar, Shiran Zada, Oran Lang, Omer Tov, Huiwen Chang, Tali Dekel, Inbar Mosseri, and Michal Irani.
\newblock Imagic: Text-based real image editing with diffusion models.
\newblock In \emph{Proceedings of the IEEE/CVF Conference on Computer Vision and Pattern Recognition}, pages 6007--6017, 2023.

\bibitem[Kim et~al.(2022)Kim, Han, and Ye]{kim2022diffusemorph}
Boah Kim, Inhwa Han, and Jong~Chul Ye.
\newblock Diffusemorph: Unsupervised deformable image registration using diffusion model.
\newblock In \emph{Computer Vision--ECCV 2022: 17th European Conference, Tel Aviv, Israel, October 23--27, 2022, Proceedings, Part XXXI}, pages 347--364. Springer, 2022.

\bibitem[Kingma and Ba(2014)]{kingma2014adam}
Diederik~P Kingma and Jimmy Ba.
\newblock Adam: A method for stochastic optimization.
\newblock \emph{arXiv preprint arXiv:1412.6980}, 2014.

\bibitem[Kingma and Welling(2013)]{kingma2013auto}
Diederik~P Kingma and Max Welling.
\newblock Auto-encoding variational bayes.
\newblock \emph{arXiv preprint arXiv:1312.6114}, 2013.

\bibitem[LaMontagne et~al.(2019)LaMontagne, Benzinger, Morris, Keefe, Hornbeck, Xiong, Grant, Hassenstab, Moulder, Vlassenko, et~al.]{lamontagne2019oasis}
Pamela~J LaMontagne, Tammie~LS Benzinger, John~C Morris, Sarah Keefe, Russ Hornbeck, Chengjie Xiong, Elizabeth Grant, Jason Hassenstab, Krista Moulder, Andrei~G Vlassenko, et~al.
\newblock Oasis-3: longitudinal neuroimaging, clinical, and cognitive dataset for normal aging and alzheimer disease.
\newblock \emph{medrxiv}, pages 2019--12, 2019.

\bibitem[Li et~al.(2024)Li, Li, and Hoi]{li2024blip}
Dongxu Li, Junnan Li, and Steven Hoi.
\newblock Blip-diffusion: Pre-trained subject representation for controllable text-to-image generation and editing.
\newblock \emph{Advances in Neural Information Processing Systems}, 36, 2024.

\bibitem[Li et~al.(2021)Li, Fan, Wang, Ma, and Cui]{li2021tackling}
Wei Li, Li~Fan, Zhenyu Wang, Chao Ma, and Xiaohui Cui.
\newblock Tackling mode collapse in multi-generator gans with orthogonal vectors.
\newblock \emph{Pattern Recognition}, 110:\penalty0 107646, 2021.

\bibitem[Liu et~al.(2021)Liu, Xing, Stone, Zhuo, Reese, Prince, El~Fakhri, and Woo]{liu2021generative}
Xiaofeng Liu, Fangxu Xing, Maureen Stone, Jiachen Zhuo, Timothy Reese, Jerry~L Prince, Georges El~Fakhri, and Jonghye Woo.
\newblock Generative self-training for cross-domain unsupervised tagged-to-cine mri synthesis.
\newblock In \emph{International Conference on Medical Image Computing and Computer-Assisted Intervention}, pages 138--148. Springer, 2021.

\bibitem[Marcus et~al.(2007)Marcus, Wang, Parker, Csernansky, Morris, and Buckner]{marcus2007open}
Daniel~S Marcus, Tracy~H Wang, Jamie Parker, John~G Csernansky, John~C Morris, and Randy~L Buckner.
\newblock Open access series of imaging studies (oasis): cross-sectional mri data in young, middle aged, nondemented, and demented older adults.
\newblock \emph{Journal of cognitive neuroscience}, 19\penalty0 (9):\penalty0 1498--1507, 2007.

\bibitem[Maz{\'e} and Ahmed(2023)]{maze2023diffusion}
Fran{\c{c}}ois Maz{\'e} and Faez Ahmed.
\newblock Diffusion models beat gans on topology optimization.
\newblock In \emph{Proceedings of the AAAI conference on artificial intelligence}, volume~37, pages 9108--9116, 2023.

\bibitem[Meng et~al.(2021)Meng, He, Song, Song, Wu, Zhu, and Ermon]{meng2021sdedit}
Chenlin Meng, Yutong He, Yang Song, Jiaming Song, Jiajun Wu, Jun-Yan Zhu, and Stefano Ermon.
\newblock Sdedit: Guided image synthesis and editing with stochastic differential equations.
\newblock \emph{arXiv:2108.01073}, 2021.

\bibitem[Miller et~al.(2002)Miller, Trouv{\'e}, and Younes]{miller2002metrics}
Michael~I Miller, Alain Trouv{\'e}, and Laurent Younes.
\newblock On the metrics and euler--lagrange equations of computational anatomy.
\newblock \emph{Annual Review of Biomedical Engineering}, 4\penalty0 (1):\penalty0 375--405, 2002.

\bibitem[Mukhopadhyay et~al.(2023)Mukhopadhyay, Gwilliam, Agarwal, Padmanabhan, Swaminathan, Hegde, Zhou, and Shrivastava]{mukhopadhyay2023diffusion}
Soumik Mukhopadhyay, Matthew Gwilliam, Vatsal Agarwal, Namitha Padmanabhan, Archana Swaminathan, Srinidhi Hegde, Tianyi Zhou, and Abhinav Shrivastava.
\newblock Diffusion models beat gans on image classification, 2023.

\bibitem[Nguyen et~al.(2024)Nguyen, Li, Ojha, and Lee]{nguyen2024visual}
Thao Nguyen, Yuheng Li, Utkarsh Ojha, and Yong~Jae Lee.
\newblock Visual instruction inversion: Image editing via image prompting.
\newblock \emph{Advances in Neural Information Processing Systems}, 36, 2024.

\bibitem[Oktay et~al.(2018)Oktay, Schlemper, Folgoc, Lee, Heinrich, Misawa, Mori, McDonagh, Hammerla, Kainz, et~al.]{oktay2018attention}
Ozan Oktay, Jo~Schlemper, Loic~Le Folgoc, Matthew Lee, Mattias Heinrich, Kazunari Misawa, Kensaku Mori, Steven McDonagh, Nils~Y Hammerla, Bernhard Kainz, et~al.
\newblock Attention u-net: Learning where to look for the pancreas.
\newblock \emph{arXiv preprint arXiv:1804.03999}, 2018.

\bibitem[{\"O}zbey et~al.(2023){\"O}zbey, Dalmaz, Dar, Bedel, {\"O}zturk, G{\"u}ng{\"o}r, and Cukur]{ozbey2023unsupervised}
Muzaffer {\"O}zbey, Onat Dalmaz, Salman~UH Dar, Hasan~A Bedel, {\c{S}}aban {\"O}zturk, Alper G{\"u}ng{\"o}r, and Tolga Cukur.
\newblock Unsupervised medical image translation with adversarial diffusion models.
\newblock \emph{IEEE Transactions on Medical Imaging}, 42\penalty0 (12):\penalty0 3524--3539, 2023.

\bibitem[Patashnik et~al.(2023)Patashnik, Garibi, Azuri, Averbuch-Elor, and Cohen-Or]{patashnik2023localizing}
Or~Patashnik, Daniel Garibi, Idan Azuri, Hadar Averbuch-Elor, and Daniel Cohen-Or.
\newblock Localizing object-level shape variations with text-to-image diffusion models.
\newblock In \emph{Proceedings of the IEEE/CVF International Conference on Computer Vision}, pages 23051--23061, 2023.

\bibitem[Radford et~al.(2021)Radford, Kim, Hallacy, Ramesh, Goh, Agarwal, Sastry, Askell, Mishkin, Clark, Krueger, and Sutskever]{radford2021learning}
Alec Radford, Jong~Wook Kim, Chris Hallacy, Aditya Ramesh, Gabriel Goh, Sandhini Agarwal, Girish Sastry, Amanda Askell, Pamela Mishkin, Jack Clark, Gretchen Krueger, and Ilya Sutskever.
\newblock Learning transferable visual models from natural language supervision, 2021.

\bibitem[Reinhold et~al.(2021)Reinhold, Carass, and Prince]{reinhold2021structural}
Jacob~C Reinhold, Aaron Carass, and Jerry~L Prince.
\newblock A structural causal model for mr images of multiple sclerosis.
\newblock In \emph{International Conference on Medical Image Computing and Computer-Assisted Intervention}, pages 782--792. Springer, 2021.

\bibitem[Ronneberger et~al.(2015)Ronneberger, Fischer, and Brox]{ronneberger2015u}
Olaf Ronneberger, Philipp Fischer, and Thomas Brox.
\newblock U-net: Convolutional networks for biomedical image segmentation.
\newblock In \emph{International Conference on Medical image computing and computer-assisted intervention}, pages 234--241. Springer, 2015.

\bibitem[Salimans et~al.(2016)Salimans, Goodfellow, Zaremba, Cheung, Radford, and Chen]{salimans2016improved}
Tim Salimans, Ian Goodfellow, Wojciech Zaremba, Vicki Cheung, Alec Radford, and Xi~Chen.
\newblock Improved techniques for training gans.
\newblock \emph{Advances in neural information processing systems}, 29, 2016.

\bibitem[Sharma et~al.(2023)Sharma, Dhall, and Subramanian]{sharma2023medic}
Gulshan Sharma, Abhinav Dhall, and Ramanathan Subramanian.
\newblock Medic: Mitigating eeg data scarcity via class-conditioned diffusion model.
\newblock In \emph{Deep Generative Models for Health Workshop NeurIPS 2023}, 2023.

\bibitem[Starck et~al.(2025)Starck, Sideri-Lampretsa, Kainz, Menten, Mueller, and Rueckert]{starck2025diff}
Sophie Starck, Vasiliki Sideri-Lampretsa, Bernhard Kainz, Martin~J Menten, Tamara~T Mueller, and Daniel Rueckert.
\newblock Diff-def: Diffusion-generated deformation fields for conditional atlases.
\newblock \emph{IEEE Transactions on Medical Imaging}, 2025.

\bibitem[Tezcan et~al.(2018)Tezcan, Baumgartner, Luechinger, Pruessmann, and Konukoglu]{tezcan2018mr}
Kerem~C Tezcan, Christian~F Baumgartner, Roger Luechinger, Klaas~P Pruessmann, and Ender Konukoglu.
\newblock Mr image reconstruction using deep density priors.
\newblock \emph{IEEE transactions on medical imaging}, 38\penalty0 (7):\penalty0 1633--1642, 2018.

\bibitem[Uchiyama et~al.(2017)Uchiyama, Sakurai, Mishima, Arita, Okayasu, Shimada, and Taniguchi]{uchiyama2017easy}
Hideaki Uchiyama, Shunsuke Sakurai, Masashi Mishima, Daisaku Arita, Takashi Okayasu, Atsushi Shimada, and Rin-ichiro Taniguchi.
\newblock An easy-to-setup 3d phenotyping platform for komatsuna dataset.
\newblock In \emph{Proceedings of the IEEE international conference on computer vision workshops}, pages 2038--2045, 2017.

\bibitem[Unterthiner et~al.(2018)Unterthiner, Van~Steenkiste, Kurach, Marinier, Michalski, and Gelly]{unterthiner2018towards}
Thomas Unterthiner, Sjoerd Van~Steenkiste, Karol Kurach, Raphael Marinier, Marcin Michalski, and Sylvain Gelly.
\newblock Towards accurate generative models of video: A new metric \& challenges.
\newblock \emph{arXiv:1812.01717}, 2018.

\bibitem[Vialard et~al.(2012)Vialard, Risser, Rueckert, and Cotter]{vialard2012}
Fran{\c{c}}ois~X Vialard, Laurent Risser, Daniel Rueckert, and Colin~J Cotter.
\newblock Diffeomorphic 3d image registration via geodesic shooting using an efficient adjoint calculation.
\newblock \emph{International Journal of Computer Vision}, 97\penalty0 (2):\penalty0 229--241, 2012.

\bibitem[Wang et~al.(2004)Wang, Bovik, Sheikh, and Simoncelli]{wang2004image}
Zhou Wang, Alan~C Bovik, Hamid~R Sheikh, and Eero~P Simoncelli.
\newblock Image quality assessment: from error visibility to structural similarity.
\newblock \emph{IEEE transactions on image processing}, 13\penalty0 (4):\penalty0 600--612, 2004.

\bibitem[Wu and Zhang(2023)]{wu2023neurepdiff}
Nian Wu and Miaomiao Zhang.
\newblock Neurepdiff: Neural operators to predict geodesics in deformation spaces.
\newblock In \emph{International Conference on Information Processing in Medical Imaging}, pages 588--600. Springer, 2023.

\bibitem[Wu and Zhang(2024)]{wu2024learning}
Nian Wu and Miaomiao Zhang.
\newblock Learning geodesics of geometric shape deformations from images.
\newblock \emph{The Journal of Machine Learning for Biomedical Imaging}, 2024.

\bibitem[Wu et~al.(2024)Wu, Xing, and Zhang]{wu2024tlrn}
Nian Wu, Jiarui Xing, and Miaomiao Zhang.
\newblock Tlrn: Temporal latent residual networks for large deformation image registration.
\newblock In \emph{International Conference on Medical Image Computing and Computer-Assisted Intervention}, pages 728--738. Springer, 2024.

\bibitem[Wu et~al.(2025)Wu, Jayakumar, Xing, and Zhang]{wu2025igg}
Nian Wu, Nivetha Jayakumar, Jiarui Xing, and Miaomiao Zhang.
\newblock Igg: Image generation informed by geodesic dynamics in deformation spaces.
\newblock In \emph{International Conference on Information Processing in Medical Imaging}, pages 232--246. Springer, 2025.

\bibitem[Xie et~al.(2025)Xie, Zhang, Weng, Zhu, and Luo]{xie2025meddiff}
Jianhao Xie, Ziang Zhang, Zhenyu Weng, Yuesheng Zhu, and Guibo Luo.
\newblock Meddiff-ft: Data-efficient diffusion model fine-tuning with structural guidance for controllable medical image synthesis.
\newblock In \emph{International Conference on Medical Image Computing and Computer-Assisted Intervention}, pages 306--316. Springer, 2025.

\bibitem[Xing et~al.(2024)Xing, Jayakumar, Wu, Wang, Epstein, and Zhang]{xing2024lamod}
Jiarui Xing, Nivetha Jayakumar, Nian Wu, Yu~Wang, Frederick~H Epstein, and Miaomiao Zhang.
\newblock Lamod: Latent motion diffusion model for myocardial strain generation.
\newblock In \emph{International Workshop on Shape in Medical Imaging}, pages 164--177. Springer, 2024.

\bibitem[Xing et~al.(2025)Xing, Xia, Zhang, Chen, Yu, Liu, Liu, Wang, Shan, and Wong]{xing2025dynamicrafter}
Jinbo Xing, Menghan Xia, Yong Zhang, Haoxin Chen, Wangbo Yu, Hanyuan Liu, Gongye Liu, Xintao Wang, Ying Shan, and Tien-Tsin Wong.
\newblock Dynamicrafter: Animating open-domain images with video diffusion priors.
\newblock In \emph{European Conference on Computer Vision}, pages 399--417. Springer, 2025.

\bibitem[Yang et~al.(2019)Yang, Dvornek, Zhang, Chapiro, Lin, and Duncan]{yang2019unsupervised}
Junlin Yang, Nicha~C Dvornek, Fan Zhang, Julius Chapiro, MingDe Lin, and James~S Duncan.
\newblock Unsupervised domain adaptation via disentangled representations: Application to cross-modality liver segmentation.
\newblock In \emph{International Conference on Medical Image Computing and Computer-Assisted Intervention}, pages 255--263. Springer, 2019.

\bibitem[Yang et~al.(2024)Yang, Teng, Zheng, Ding, Huang, Xu, Yang, Hong, Zhang, Feng, et~al.]{yang2024cogvideox}
Zhuoyi Yang, Jiayan Teng, Wendi Zheng, Ming Ding, Shiyu Huang, Jiazheng Xu, Yuanming Yang, Wenyi Hong, Xiaohan Zhang, Guanyu Feng, et~al.
\newblock Cogvideox: Text-to-video diffusion models with an expert transformer.
\newblock \emph{arXiv:2408.06072}, 2024.

\bibitem[Yoon et~al.(2023)Yoon, Zhang, Suk, Guo, and Li]{yoon2023sadm}
Jee~Seok Yoon, Chenghao Zhang, Heung-Il Suk, Jia Guo, and Xiaoxiao Li.
\newblock Sadm: Sequence-aware diffusion model for longitudinal medical image generation.
\newblock In \emph{International Conference on Information Processing in Medical Imaging}, pages 388--400. Springer, 2023.

\bibitem[Younes et~al.(2009)Younes, Arrate, and Miller]{younes2009evolutions}
Laurent Younes, Felipe Arrate, and Michael~I Miller.
\newblock Evolutions equations in computational anatomy.
\newblock \emph{NeuroImage}, 45\penalty0 (1):\penalty0 S40--S50, 2009.

\bibitem[Yu et~al.(2025)Yu, Dou, Long, Lin, Li, Liu, M{\"u}ller, Komura, Habermann, Theobalt, et~al.]{yu2025surf}
Zhengming Yu, Zhiyang Dou, Xiaoxiao Long, Cheng Lin, Zekun Li, Yuan Liu, Norman M{\"u}ller, Taku Komura, Marc Habermann, Christian Theobalt, et~al.
\newblock Surf-d: Generating high-quality surfaces of arbitrary topologies using diffusion models.
\newblock In \emph{European Conference on Computer Vision}, pages 419--438. Springer, 2025.

\bibitem[Zhang et~al.(2024)Zhang, Chen, Huang, Zhu, Ding, and Shen]{zhang2024development}
Kai Zhang, Geng Chen, Shijie Huang, Fangmei Zhu, Zhongxiang Ding, and Dinggang Shen.
\newblock Development-driven diffusion model for longitudinal prediction of fetal brain mri with unpaired data.
\newblock \emph{IEEE Transactions on Medical Imaging}, 2024.

\bibitem[Zhang et~al.(2017)Zhang, Liao, Dalca, Turk, Luo, Grant, and Golland]{zhang2017frequency}
Miaomiao Zhang, Ruizhi Liao, Adrian~V Dalca, Esra~A Turk, Jie Luo, P~Ellen Grant, and Polina Golland.
\newblock Frequency diffeomorphisms for efficient image registration.
\newblock In \emph{International Conference on Information Processing in Medical Imaging}, pages 559--570. Springer, 2017.

\bibitem[Zhang et~al.(2018)Zhang, Li, and Yu]{zhang2018convergence}
Zhaoyu Zhang, Mengyan Li, and Jun Yu.
\newblock On the convergence and mode collapse of gan.
\newblock In \emph{SIGGRAPH Asia 2018 Technical Briefs}, pages 1--4. 2018.

\end{thebibliography}


\end{document}